\documentclass[10pt,twocolumn,letterpaper]{article}

\usepackage{cvpr}
\usepackage{times}
\usepackage{epsfig}
\usepackage{graphicx}
\usepackage{amsmath}
\usepackage{amssymb}

\usepackage{multirow}
\usepackage{booktabs}
\usepackage{float}
\usepackage{caption}
\usepackage[dvipsnames]{xcolor}
\usepackage{tabularx}

\def\cq{\textcolor{black}}

\usepackage{hyperref}
\hypersetup{colorlinks,allcolors=green}

\cvprfinalcopy 

\def\cvprPaperID{4253} 

\begin{document}


\title{\vspace{-20pt} Intelligent Home 3D: Automatic 3D-House Design  \\from Linguistic Descriptions Only}

\newcommand{\sexyname}{HPGM\xspace}

\author{Qi Chen$^{1,2}$\thanks{Authors contributed equally.}
	\and
	Qi Wu$^{3*}$
	\and
	Rui Tang$^4$
	\and
	Yuhan Wang$^4$
	\and
	Shuai Wang$^4$
	\and
	Mingkui Tan$^{1}$\thanks{Corresponding author.}
	\and
	$^1$South China University of Technology, 
	$^2$Guangzhou  Laboratory,  Guangzhou, China \\
	$^3$Australian Centre for Robotic Vision, University of Adelaide, Australia\\
	$^4$Kujiale, Inc.
	\and
	{\tt\small sechenqi@mail.scut.edu.cn} \\
	{\tt\small \{ati,daishu,luorui\}@qunhemail.com}
	\and
	{\tt\small qi.wu01@adelaide.edu.au} \\
	{\tt\small mingkuitan@scut.edu.cn}
}




\maketitle

\begin{abstract}
	Home design is a complex task that normally requires architects to finish with their professional skills and tools. It will be fascinating that if one can produce a house plan intuitively without knowing much knowledge about home design and experience of using  complex designing tools, for example, via natural language. In this paper, we formulate it as a language conditioned visual content generation problem that is further divided into a floor plan generation and an interior texture (such as floor and wall) synthesis task. The only control signal of the generation process is the linguistic expression given by users that describe the house details. To this end, we propose a House Plan Generative Model (HPGM) that first translates the language input to a structural graph representation and then predicts the layout of rooms with a Graph Conditioned Layout Prediction Network (GC-LPN) and generates the interior texture with a Language Conditioned Texture GAN (LCT-GAN). With some post-processing, the final product of this task is a 3D house model. To train and evaluate our model, we build the first Text--to--3D House Model dataset. 
\end{abstract}



\section{Introduction}

Everyone wants a dream home, but not everyone can design home by themselves. Home design is a complex task that is normally done by certificated architects, who have to receive several years of training on designing, planning and using special designing tools. To design a home, they typically start by collecting a list of requirements for a building layout.
Then, they use trial-and-error to generate layouts with a combination of intuition and prior experience. This usually takes from a couple of days to several weeks and has high requirements for professional knowledge. 

\begin{figure}[t]
	\centering
	\includegraphics[width=0.95\linewidth]{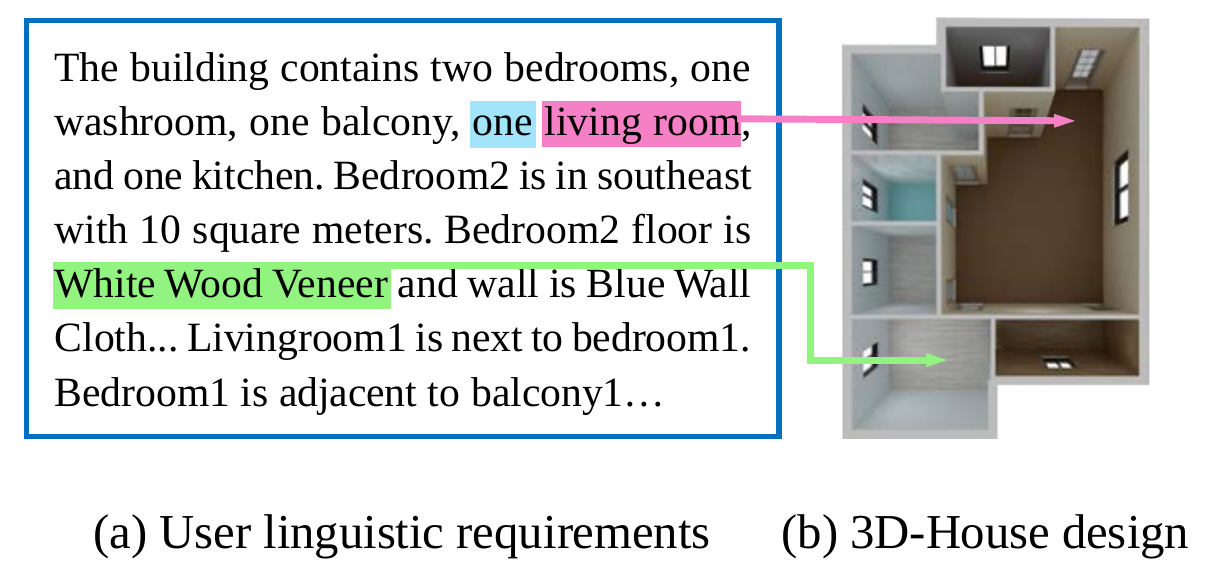}
	\vspace{-3pt}
	\caption{An example of generated 3D house with description using HPGM on the Text--to--3D House Model dataset. 3D-house generation from requirement seeks to design a 3D building automatically from given linguistic descriptions.
	}
	\vspace{-15pt}
	\label{title}
\end{figure}

It will be fantastic if we can design our own home by ourselves. We may not have home design knowledge and have no idea how to use those complicated professional designing tools, but we have 
strong linguistic ability 
to express our interests and desire. Thus, for time-saving and allowing people without expertise to participate in the design, we propose to use linguistic expressions as the guidance to generate home design plans, as shown in Figure \ref{title}. 
Thanks to the fast development of Deep Learning (DL)~\cite{chen2019relation,guo2020closed,guo2019nat,zeng2019breaking,zeng2019graph,zhang2019whole,zhang2019collaborative,zhao2018adaptive,zhuang2018discrimination}, especially Generative Adversarial Network (GAN)~\cite{pmlr-v80-cao18a,cao2019multi,goodfellow2014generative,guo2019auto} and vision-language research~\cite{huang2020graph,zeng2020dense}, we can turn this problem into a text-to-image generation problem, which has been studied in~\cite{li2019object, qiao2019mirrorgan, reed2016generative, xu2018attngan, yin2019semantics}.
However, it is non-trivial to directly apply these methods on our new task because there exist two new technical challenges: \textbf{1)} A floor plan is a structured layout which pays more attention to the correctness of size, direction, and connection of different blocks, while the conventional text-to-image task focuses more on pixel-level generation accuracy.
\textbf{2)} The interior texture such as floor and wall needs neater and more stable pixel generation than general images and should be well aligned with the given descriptions.

To tackle the above issues, we propose a House Plan Generative Model (HPGM) to generate home plans from given linguistic descriptions. The HPGM first uses a Stanford Scene Graph Parser~\cite{schuster2015generating} to parse the language to a structural graph layout, where nodes represent room types associated with size, room floor (wall) colour and material. Edges between nodes indicate whether rooms are connected or not. We then divide the house plan generation process into two sub-tasks: \textit{building layout generation} and \textit{texture synthesis}. Both of them are conditioned on the above extracted structural graph. Specifically, we design a Graph Conditioned Layout Prediction Network (GC-LPN) which applies a Graph Convolutional Network~\cite{kipf2016semi} to encode the graph as a feature representation and predicts the room layouts via bounding box regressions. The predicted room layouts are sent to a floor plan post-processing step, which outputs a featured floor plan with doors, windows, walls, \etc. To generate floor and wall textures, we design a Language Conditioned Texture GAN  (LCT-GAN) that takes the encoded 
text representations as input and generates texture images with three designed adversarial, material-aware, and colour-aware losses. The generated floor plan and texture images
are sent to an auto 3D rendering system to produce the final rendered 3D house plan.

For 3D house generation from linguistic description,
we build the first Text–to–3D House Model dataset that 
contains a 2D floor plan and two texture (floor and wall) patches for each room in the house. We evaluate the room layout generation and texture generation ability of our model separately. The room layout accuracy is evaluated based on the IoU (Intersection over Union) between the predicted room bounding boxes and the ground-truth annotation. The generated interior textures are evaluated with popular image generation metrics such as \textit{Fr\'echet Inception Distance (FID)}~\cite{heusel2017gans} and \textit{Multi-scale Structural Similarity  (MS-SSIM)}~\cite{wang2003multiscale}. Our proposed GC-LPN and LCT-GAN outperform the baseline and state-of-the-art models in a large margin.
Besides, a generalisation ability evaluation of our LCT-GAN is carried out.
We also perform a human evaluation on our final products -- 3D house plans, which shows that $39.41\%$ pass it. 

We highlight our principal contributions as follows:

\begin{itemize}
	\vspace{-6pt}
	\item We propose a novel architecture, called House Plan Generative Model (HPGM), which is able to generate 3D house models with given linguistic expressions. To reduce the difficulty, we divide the generation task into two sub-tasks to generate floor plans and interior textures, separately.
	\vspace{-6pt}
	\item To achieve the goal of synthesising 3D building model from the text, we collect a new dataset consisting of the building layouts, texture images, and their corresponding natural language expressions.
	\vspace{-6pt}
	\item Extensive experiments show the effectiveness of our proposed method on both qualitative and quantitative metrics. We also study the generalisation ability of the proposed method by generating unseen data with the given new texts.
\end{itemize}

\section{Related Work}
\vspace{-5pt}
\paragraph{Building layout design.}
Several existing methods have been proposed for generating building layouts automatically~\cite{bao2013generating, chaillou2019archigan, merrell2010computer, peng2014computing, wu2018miqp}. 
However, most of these methods generate the building layouts by merely adjusting the interior edges in a given building outline.
Specifically, Merrel \etal~\cite{merrell2010computer} generate residential building layouts using a Bayesian network trained in architectural programs.
Based on an initial layout, Bao \etal~\cite{bao2013generating} formulate a constrained optimisation to characterise the local shape spaces and then link them to a portal graph to obtain the objective layout.
Peng \etal~\cite{peng2014computing} devise a framework to yield the floor plan by tiling an arbitrarily shaped building outline with a set of deformable templates.
Wu \etal~\cite{wu2018miqp} develop a framework that generates building interiors with high-level requirements. More recently, Wu \etal~\cite{Wu_DeepLayout_2019} propose a data-driven floor plan generating system by learning thousands of samples. However, the above methods require either a given building outline or a detailed structured representation as the input while we generate the room layouts with human verbal commands.

\vspace{-12pt}
\paragraph{Texture synthesis.}
Many existing works in terms of texture generation focus on transferring a given image into a new texture style~\cite{johnson2016perceptual, li2016precomputed, ulyanov2017improved} or synthesising a new texture image based on the input texture~\cite{gatys2015texture, tesfaldet2018two, xian2018texturegan}. 
Different from that, we aim to solve the problem that generates texture images with given linguistic expressions. 
The closest alternative to our task is texture generation from random noise~\cite{bergmann2017learning, jetchev2016texture}.
Specifically, Jetchev \etal~\cite{jetchev2016texture} propose a texture synthesis method based on GANs, which can learn a generating process from the given example images. Recently, to obtain more impressive images, Bergmann \etal~\cite{bergmann2017learning} incorporate the periodical information into the generative model, which makes the model have the ability to synthesise periodic texture seamlessly.
Even if these methods have a strong ability to produce plausible images, they have limited real-world applications due to the uncontrollable and randomly generated results. We use natural language as the control signal for texture generation.

\begin{figure*}[!t]
	\vspace{-10pt}
	\begin{center}
		\includegraphics[width=1.0\linewidth]{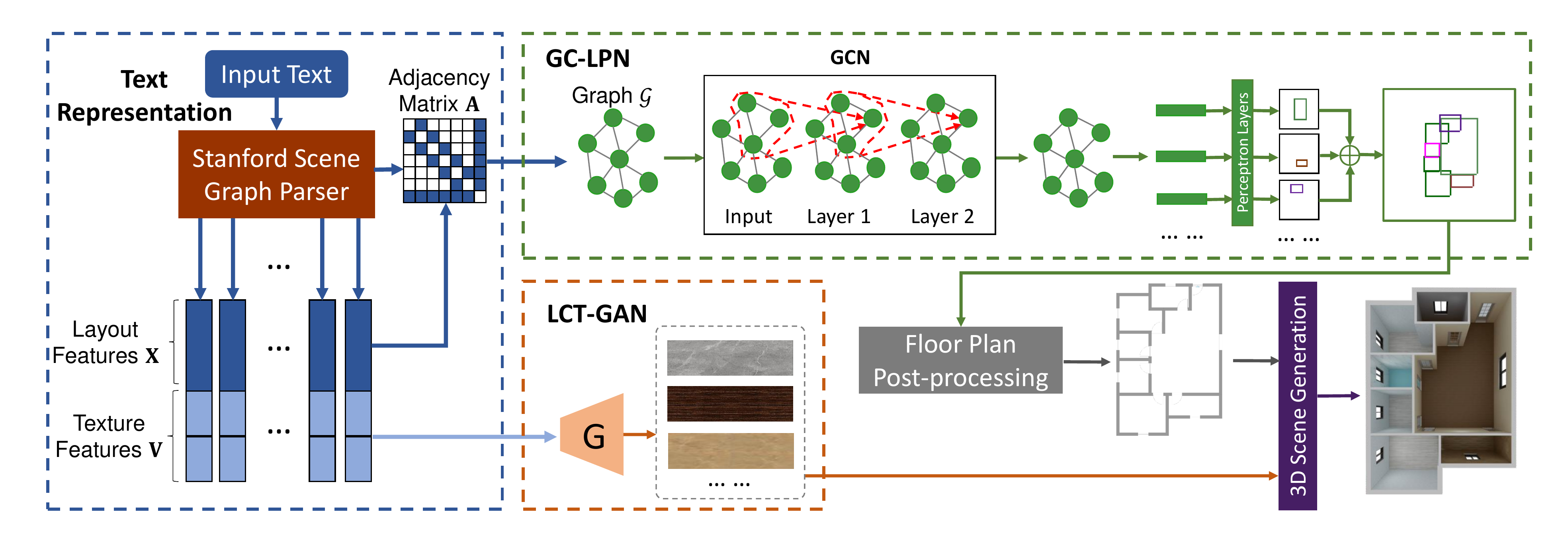}
	\end{center}
	\vspace{-25pt}
	\caption{The overview of HPGM. We use Stanford Scene Graph Parser to parse given textual input and obtain the structural text representations $\mathbf{X}$, $\mathbf{V}$ and $\mathbf{A}$.
		Based on $\mathbf{X}$ and $\mathbf{A}$, GC-LPN yields a rough building layout using a graph convolutional network, followed by post-processing to refine the generated floor plan. LCT-GAN synthesises the corresponding textures for each room according to $\mathbf{V}$. Last, a 3D scene generation method is used to produce the objective 3D house plan.
	}
	\vspace{-10pt}
	\label{fig:architecture}
\end{figure*}

\vspace{-12pt}
\paragraph{Text to image generation.}
For generating an image from text, many GAN-based methods~\cite{johnson2018image, li2019storygan, qiao2019mirrorgan, reed2016generative, xu2018attngan, yin2019semantics, Han17stackgan2, zhang2017stackgan, zhu2019dm} have been proposed in this area. Reed \etal~\cite{reed2016generative} transform the given sentence into text embedding and then generate image conditioning on the extracted embedding. Furthermore, to yield more realistic images, Zhang \etal~\cite{zhang2017stackgan} propose a hierarchical network, called StackGAN, which generates images with different sizes (from coarse to fine). Meanwhile, they introduce a conditioning augmentation method to avoid the discontinuity in the latent manifold of text embedding. Based on StackGAN, Xu \etal~\cite{xu2018attngan} develop an attention mechanism, which ensures the alignment between generated fine-grained images and the corresponding word-level conditions. More recently, to preserve the semantic consistency, Qiao \etal~\cite{qiao2019mirrorgan} consider both text-to-image and image-to-text problems jointly.

\vspace{-3pt}
\section{Proposed Method}
\vspace{-3pt}

In this paper, we focus on 3D-house generation from requirements, which seeks to design a 3D building automatically conditioned on the given linguistic descriptions.
Due to the intrinsic complexity of 3D-house design, we divide the generation process into two sub-tasks: \textit{building layout generation} and \textit{texture synthesis}, which produce floor plan and corresponding room features (\ie, textures of each room), respectively.

To complete the above two tasks, we propose a House Plan Generative Model (HPGM) to automatically generate a 3D home design  conditioned on given descriptions. As shown in Figure~\ref{fig:architecture}, the proposed HPGM consists of five components: 1) text representation block, 2) graph conditioned layout prediction network (GC-LPN), 3) floor plan post-processing, 4) language conditioned texture GAN (LCT-GAN), and  5) 3D scene generation and rendering.

In Figure~\ref{fig:architecture}, the text representation is to capture the structural text information from given texts using a Stanford Scene Graph Parser~\cite{schuster2015generating}. Based on the text representations,
GC-LPN is devised to produce a coarse building layout.   To obtain a real-world 2D floor plan, we send the generated layout to a {floor plan post-processing} step to refine the coarse building layout to yield a floor plan with windows and doors.
To synthesise the interior textures of each room, we further devise a Language Conditioned Texture GAN (LCT-GAN) to yield the controllable and neat images according to the semantic text representations. Last, we feed the generated floor plan with room features into a 3D rendering system for {3D scene generation and rendering}. The details of each component are depicted below.

\subsection{Text Representation}\label{text_representation}

The linguistic descriptions of the building include the description of the number of rooms and room types, followed by the connections between rooms, and the designing patterns of each room. Although it follows a weakly structural format, directly
using the template-based language parser is impractical due to the diversity of the linguistic descriptions.
Instead, we employ the Stanford Scene Graph Parser~\cite{schuster2015generating} with some post-processing and merging to parse the linguistic descriptions
to a structural graph 
format. For such a constructed graph, each node is a room with some properties (\eg, the room type, size, interior textures).
The edge between nodes indicates the connectivity of two rooms. 
More details of the scene graph parser can be found in 
the supplementary materials.




\cq{We use different representation as inputs in building layout generation and texture synthesis, since these two tasks require different semantic information.}
In building layout generation, we define input vectors as $\mathbf{X}\in\mathbb{R}^{N\times D}$, where $N$ refers to the number of nodes (\ie, rooms) in each layout and $D$ denotes the feature dimension. 
Each node feature $\mathbf{x}_i=\left\{\boldsymbol{\alpha}_i, \boldsymbol{\beta}_i, \boldsymbol{\gamma}_i\right\} \in \mathbb{R}^{D}$ is a triplet, where $\boldsymbol{\alpha}_i$ is the type of room (\eg, bedroom), $\boldsymbol{\beta}_i$ is the size (\eg, 20 squares) and $\boldsymbol{\gamma}_i$ is the position (\eg, southwest). All features are encoded as one-hot vectors except the size is a real value.
Moreover, to exploit the topological information elaborately, following~\cite{kipf2016semi}, we convert the input features $\mathbf{X}$ to an undirected graph $\mathcal{G}$ via introducing an adjacency matrix $\mathbf{A}\in\mathbb{R}^{N\times N}$.




In the texture synthesis task, for a given text, we transform the linguistic expression to a collection of vectors $\mathbf{V}\in\mathbb{R}^{2N\times M}$, where $2N$ refers to the number of textures in each layout and $M$ denotes the dimension of each feature vector. 
For $\mathbf{v}_i\in\mathbb{R}^{M}$, we design $\mathbf{v}_i=\left\{\mathbf{p}_i, \mathbf{q}_i\right\}$, where  $\mathbf{p}_i$ indicates the material (\eg, log, mosaic or stone brick) and $\mathbf{q}_i$ refers to the colour. 
We pre-build a 
material and colour
word vocabulary from training data so that we can classify the parsed attributes into the material or colour set.

\subsection{Graph Conditioned Layout Prediction Network}

To generate the building layouts satisfying the requirements, we propose a Graph Conditioned Layout Prediction Network (GC-LPN). We incorporate the adjacent information into the extracted features via a GCN, which facilitates the performance when generating the objective layouts. 


\begin{figure}[t]
	\begin{center}
		\includegraphics[width=0.9\linewidth]{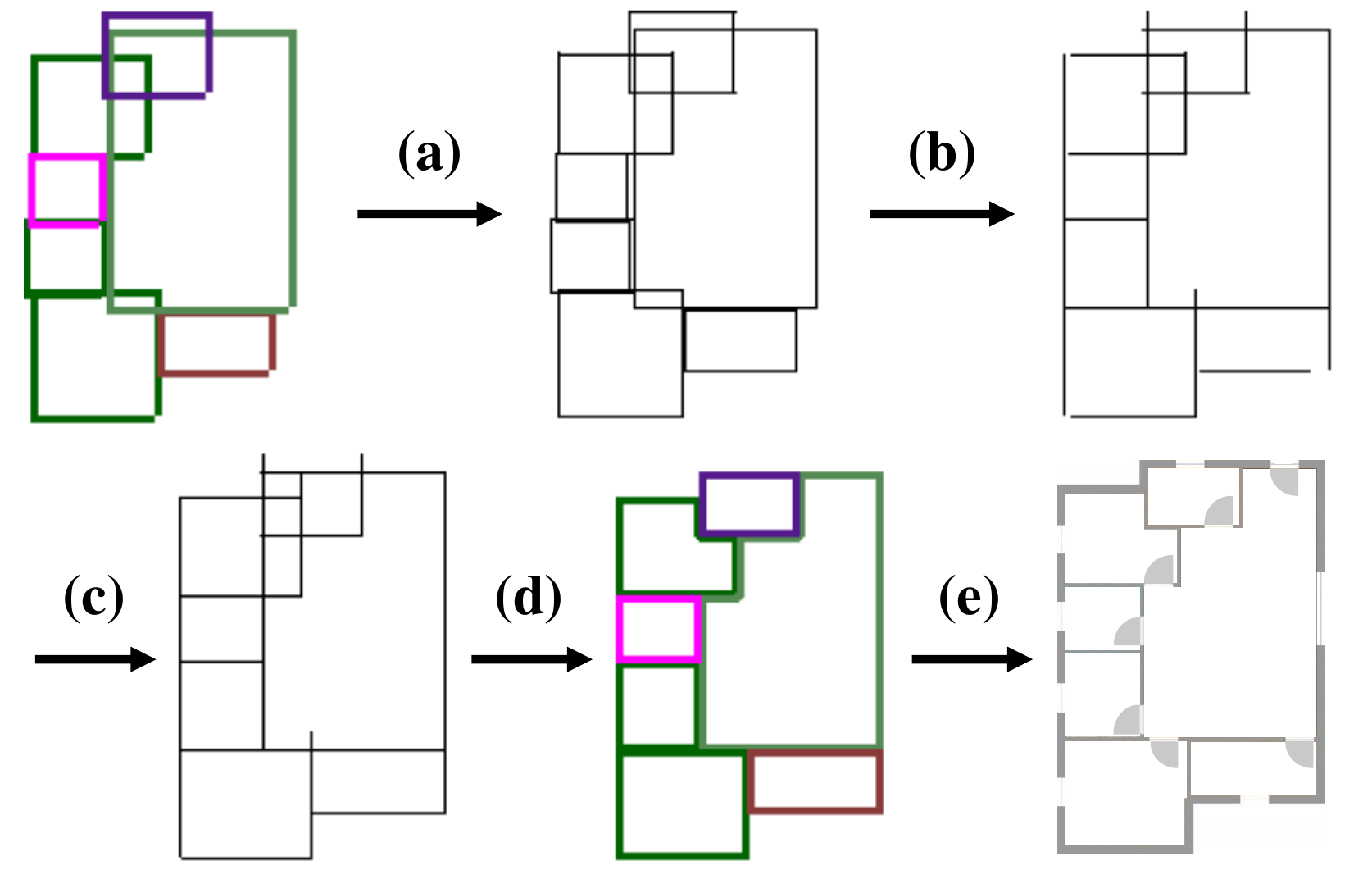}
	\end{center}
	\vspace{-15pt}
	\caption{The procedures of floor plan post-processing.}
	\vspace{-10pt}
	\label{fig:post_processing}
\end{figure}

\vspace{-9pt}
\paragraph{Graph convolutional network.}
In order to process the aforementioned graphs in an end-to-end manner, we use a graph convolutional network composed of two graph convolutional layers. Specifically, we take the feature matrix $\mathbf{X}\in\mathbb{R}^{N\times D}$ as inputs and produce a new feature matrix, where each output vector is an aggregation of a local neighbourhood of its corresponding input vector. In this way, we obtain a new feature matrix, which introduces the information across local neighbourhoods of the inputs. Note that, since we only focus on generating the layouts of resident building, the order and size of corresponding graph are small. Therefore, it is sufficient to leverage a two-layer GCN model (as shown in Figure~\ref{fig:architecture}) when introducing the information of adjacent rooms. Mathematically, 
we have
{
	\vspace{-1pt}
	\begin{equation}
	\mathbf{Y} = g(\mathbf{X}, \mathbf{A}) = \mathrm{Softmax}\left(\mathbf{A} \mathrm{ReLU}\left(\mathbf{A}\mathbf{X}\mathbf{W}_0\right)\mathbf{W}_1\right),
	\vspace{-1pt}
	\end{equation}
}where $\mathbf{W}_0 \in \mathbb{R}^{D\times D}$ and $\mathbf{W}_1 \in \mathbb{R}^{D\times D}$ 
are the weights of two graph convolutional layers.
Note that the adjacency matrix $\mathbf{A}$ only contains 1 and 0, which indicates whether pairs of nodes (rooms) are adjacent or not.
$\mathbf{Y}\in\mathbb{R}^{N\times D}$ is the structured feature. 
Then, we add the extracted feature $\mathbf{Y}$ with the input feature $\mathbf{X}$ to get the feature $\mathbf{S}\in\mathbb{R}^{N\times D}$
{
	\vspace{-3pt}
	\begin{equation}
	\mathbf{S} = \mathbf{X}\oplus\mathbf{Y},
	\vspace{-3pt}
	\end{equation}}where ``$\oplus$'' is the element-wise addition.

\vspace{-9pt}
\paragraph{Bounding box regression.}
After reasoning on the graph with GCNs, we gain a set of embedding vectors, where each vector aggregates the information across the adjacent rooms.
In order to produce the building layout, we must transform these vectors from the graph domain to the image domain. Thus, we define each room as a coarse 2D bounding box, which can be represented as $\mathbf{b}_i=(x_0, y_0, x_1, y_1)$. In this way, we cast the problem to that bounding box generation from given room embedding vectors.

In practice, we first feed the well-designed feature $\mathbf{S}$ into a two-layer perceptron network $h(\cdot)$ and predict the corresponding bounding box of each node $\mathbf{\hat{b}}_i=h(\mathbf{S}_i)=(\hat{x}_0, \hat{y}_0, \hat{x}_1, \hat{y}_1)$. 
Then, we integrate all the predicted boxes and obtain the corresponding building layout.
For training the proposed model, we minimise the objective function
{
	\vspace{-5pt}
	\begin{equation}
	\mathcal{L}_B = \frac{1}{N}\sum_{i=1}^{N} {\|\mathbf{\hat{b}}_i-\mathbf{b}_i\|}^2_2,
	\vspace{-5pt}
	\end{equation}
}where $\mathbf{b}_i$ is the ground-truth bounding box for $i^{th}$ node (\ie, the bounding box that covers the room). 

\subsection{Floor Plan Post-processing}


{\renewcommand\baselinestretch{1.0}
	\selectfont
	To transform the bounding box layout to a real-world 2D floor plan, we propose a floor plan post-processing (shown in Figure~\ref{fig:post_processing}), which consists of five steps, \ie, (a)$\sim$(e). To be specific, in Step \textbf{(a)}, we first extract boundary lines of all generated bounding boxes and then merge the adjacent segments together in Step \textbf{(b)}.
	In Step \textbf{(c)}, we further align the line segments with each other to obtain the closed polygon.
	In Step \textbf{(d)}, we judge the belonging of each closed polygon based on a weight function:
	\par}
{
	\small
	\vspace{-10pt}
	\begin{equation}
	W_{ij} = \iint \frac{1}{w_ih_i}\mathrm{exp}\left(-(\frac{x_j-c_{x_i}}{w_i})^2-(\frac{y_j-c_{y_i}}{h_i}) ^2\right) dx_j dy_j,
	\label{eq:weight_function}
	\end{equation}}
where $i=1, 2, \ldots, n$ is the $i^{th}$ original box (room) while $j=1, 2, \ldots, m$ is the $j^{th}$ aligned polygon. $W_{ij}$ denotes the weight of $j^{th}$ polygon belonging to room $i$. $c_{x_i}$ and $c_{y_i}$ indicate the central position while $w_i$ and $h_i$ are the half width and height of the $i^{th}$ bounding box. $x_j$ and $y_j$ are the coordinates in the aligned polygon.
We assign the $j^{th}$ polygon with the room type, according to the corresponding original bounding box, which has maximum weight $W$.

Finally, in Step \textbf{(e)}, we apply a simple rule-based method to add doors and windows in rooms.
Specifically,
a door or open wall is added between the living room and any other room. We set the window on the longest wall of each room and set the entrance on the wall of the biggest living room. We find these rules work in most cases and good enough to set reasonable positions, but a learning-based method may improve this process and we leave it as the future work.

\subsection{Language Conditioned Texture GAN}

\begin{figure}[t]
	\vspace{-5pt}
	\begin{center}
		\includegraphics[width=1.0\linewidth]{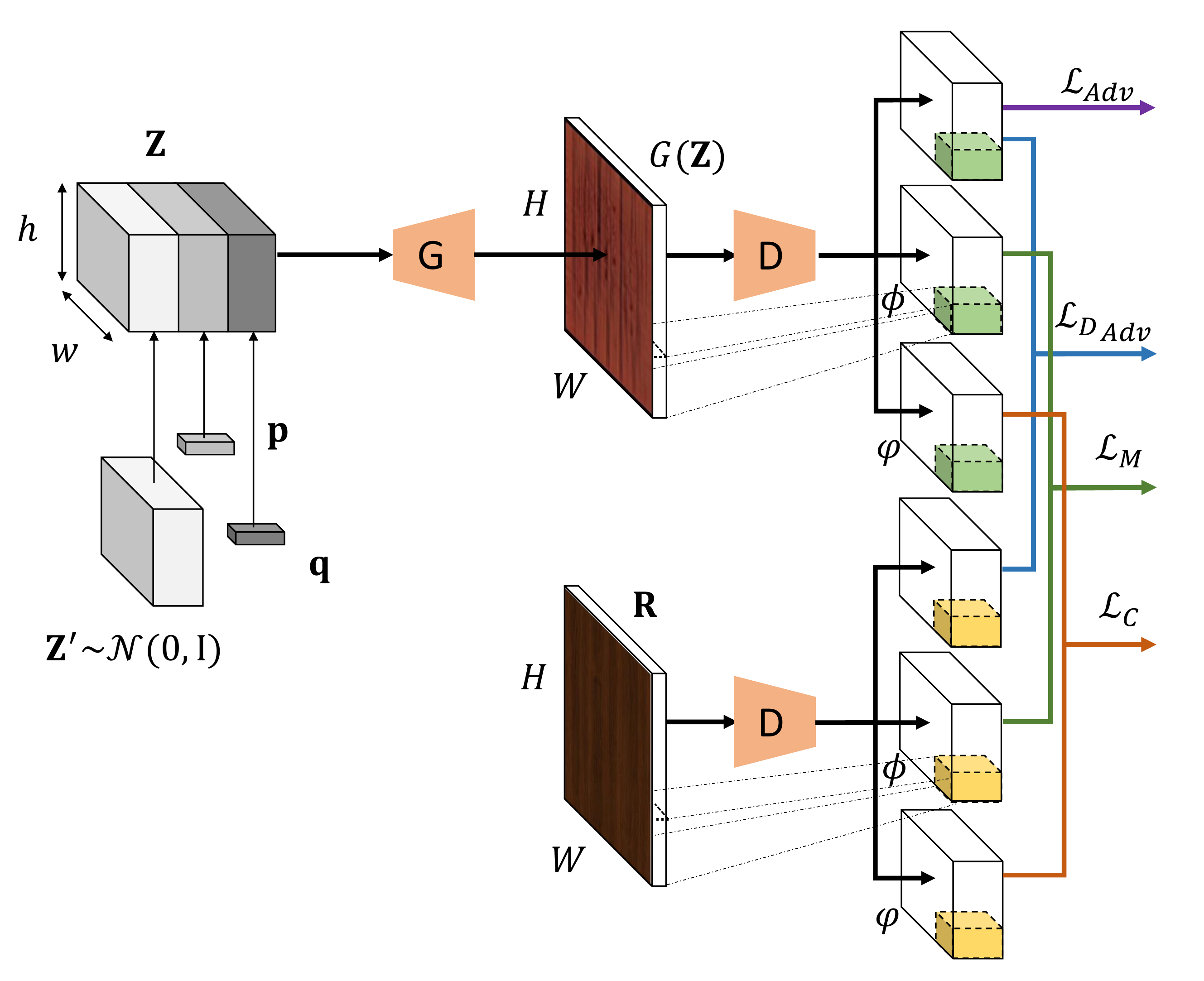}
	\end{center}
	\vspace{-20pt}
	\caption{Architecture of LCT-GAN. Generator $G$ transforms a conditional noise $\mathbf{Z}$ into an RGB image $G(\mathbf{Z})$ with fully convolutional layers. Discriminator $D$, which is used to distinguish fake data from real ones, is fed either fake image $G(\mathbf{Z})$ or real image $\mathbf{R}$. Two classifiers ($\phi$ and $\varphi$) have been added on top of $D$, which are used to impose the image into the right material and colour categories, respectively.}
	\vspace{-10pt}
	\label{fig:texture_generation}
\end{figure}

For better controlling the details of textures, we consider the texture images in terms of two fields, \ie, material and colour. 
In this way, we design a Language Conditioned Texture GAN (LCT-GAN) (in Figure~\ref{fig:texture_generation}), which can generate texture images that align with the given expressions.


\vspace{-8pt}
\paragraph{Texture generator.}

We first obtain the input noise $\mathbf{Z}^{'}\in\mathbb{R}^{w\times h\times d_1}$ from Gaussian distribution $\mathcal{N}(\mathbf{0}, \mathbf{I})$. After that, to incorporate the conditional information, we extend the aforementioned material and colour vectors $\mathbf{p}\in\mathbb{R}^{1\times 1\times d_2}$ and $\mathbf{q}\in\mathbb{R}^{1\times 1\times d_3}$ as the same size with the noise $\mathbf{Z}^{'}$ and then concatenate them together to obtain the objective input 
$\mathbf{Z}\in\mathbb{R}^{w\times h\times (d_1+d_2+d_3)}$.

Conditioning on the input tensor $\mathbf{Z}$, we generate the corresponding texture image by $G(\mathbf{Z}) \in \mathbb{R}^{W\times H\times 3}$, where $W$ and $H$ denote the width and height of the generated image, respectively.
Note that, in order to generate arbitrary size of texture, we design the generator $G$ with a fully convolutional network (FCN), which allows input $\mathbf{Z}$ with various sizes when inferring.
In practice, we establish our FCN model with only five blocks, where each block consists of a $2\times$ upsampling interpolation, a convolutional layer, a batch normalisation~\cite{ioffe2015batch} and an activation function. Due to the page limit, we put more details in the supplementary.

On the other hand, to generate texture from an expression, the generator $G$ must: 1) ensure the generated images are natural and realistic ; and 2) preserve the semantic alignment between given texts and texture images. 
To satisfy the above requirements, we propose an optimisation mechanism consisting of three losses $\mathcal{L}_{Adv}$, $\mathcal{L}_{M}$ and $\mathcal{L}_{C}$, which indicate the adversarial loss, material-aware loss and colour-aware loss, respectively.
Overall, the final objective function of the texture generator $G$ is
\vspace{-5pt}
\begin{equation}
\mathcal{L}_G = \mathcal{L}_{Adv} + \lambda_1 \mathcal{L}_{M} + \lambda_2 \mathcal{L}_{C},
\vspace{-5pt}
\end{equation}
where $\lambda_1$ and $\lambda_2$ are trade-off parameters. In experiments, we set $\lambda_1$ and $\lambda_2$ to 1 by default. We will elaborate on the modules that lead to these losses in the following sections.


\vspace{-10pt}
\paragraph{Adversarial loss.}

To synthesise the natural images, we follow the traditional GAN~\cite{goodfellow2014generative}, where the generator $G$ and discriminator $D$ compete in a two-player minimax game. Specifically, the generator $G$ tries to fool the discriminator $D$ while $D$ tries to distinguish whether the given image is real or fake/generated. Based on that, for our task, when optimising the discriminator $D$, we minimise the loss
{
	\vspace{-1pt}
	\begin{equation}
	\mathcal{L}_{D_{Adv}} = -\mathbb{E}_{\mathbf{R}\sim P_r}[\log D(\mathbf{R})] - \mathbb{E}_{\mathbf{Z}^{'}\sim P_z}[\log (1-D(G(\mathbf{Z})))],
	\vspace{-1pt}
	\end{equation}}where $P_r$ and $P_z$ denote the distributions of real samples and noise, respectively. $\mathbf{Z}$ refers to the input of $G$, as mentioned before, consisting of noise $\mathbf{Z}^{'}$ and conditions $\mathbf{p}$ and $\mathbf{q}$. On the other hand, when optimising network $G$, we use
{
	\vspace{-5pt}
	\begin{equation}
	\mathcal{L}_{Adv} = - \mathbb{E}_{\mathbf{Z}^{'}\sim P_z}[\log D(G(\mathbf{Z}))].
	\vspace{-1pt}
	\end{equation}}

\vspace{-28pt}
\paragraph{Material-aware loss.}
To preserve the semantic alignment between generated textures and given texts, we propose a material-aware loss, which is sensitive to fine-grained material categories.
To be specific, as mentioned in Section~\ref{text_representation}, we transform the linguistic descriptions to a structural format, which includes a label for each node to indicate its floor/wall material categories. 
We then add a material classifier on top of $D$, called $\phi$, which imposes the generated texture into the right category.
In this way, we obtain the posterior probability 
$\phi(\textbf{c}_m|\cdot)$
of each entry image, where $\textbf{c}_m$ refers to the category of material. Thus, we minimise the training loss of $G$ and $D$ as
{
	\vspace{-5pt}
	\begin{equation}
	\mathcal{L}_{M} = 
	- \mathbb{E}_{\mathbf{R}\sim P_r}[\log  \phi(\textbf{c}_m|\mathbf{R})]
	- \mathbb{E}_{\mathbf{Z}^{'}\sim P_z}[\log  \phi(\textbf{c}_m|G(\mathbf{Z}))].
	\vspace{-1pt}
	\end{equation}}

\vspace{-25pt}
\paragraph{Colour-aware loss.}
Similar to the above material-aware loss, instead of focusing on materials, colour-aware loss pays more attention to colour categories.
Based on the given expressions of texture colour, we cast the colour alignment as a classification problem. Specifically, we reuse the discriminator $D$ as the feature extractor and replace the last layer to be a colour classifier $\varphi$. Then, in both $G$ and $D$, we try to minimise the loss
{
	\vspace{-5pt}
	\begin{equation}
	\mathcal{L}_{C} = - \mathbb{E}_{\mathbf{R}\sim P_r}[\log  \varphi(\textbf{c}_c|\mathbf{R})] - \mathbb{E}_{\mathbf{Z}^{'}\sim P_z}[\log  \varphi(\textbf{c}_c|G(\mathbf{Z}))],
	\vspace{-1pt}
	\end{equation}}where $\varphi(\textbf{c}_c|\mathbf{R})$ is the posterior probability conditioning on the given texture image $\mathbf{R}$.

\subsection{3D Scene Generation and Rendering}
\begin{figure}[t]
	\begin{center}
		\includegraphics[width=1.0\linewidth]{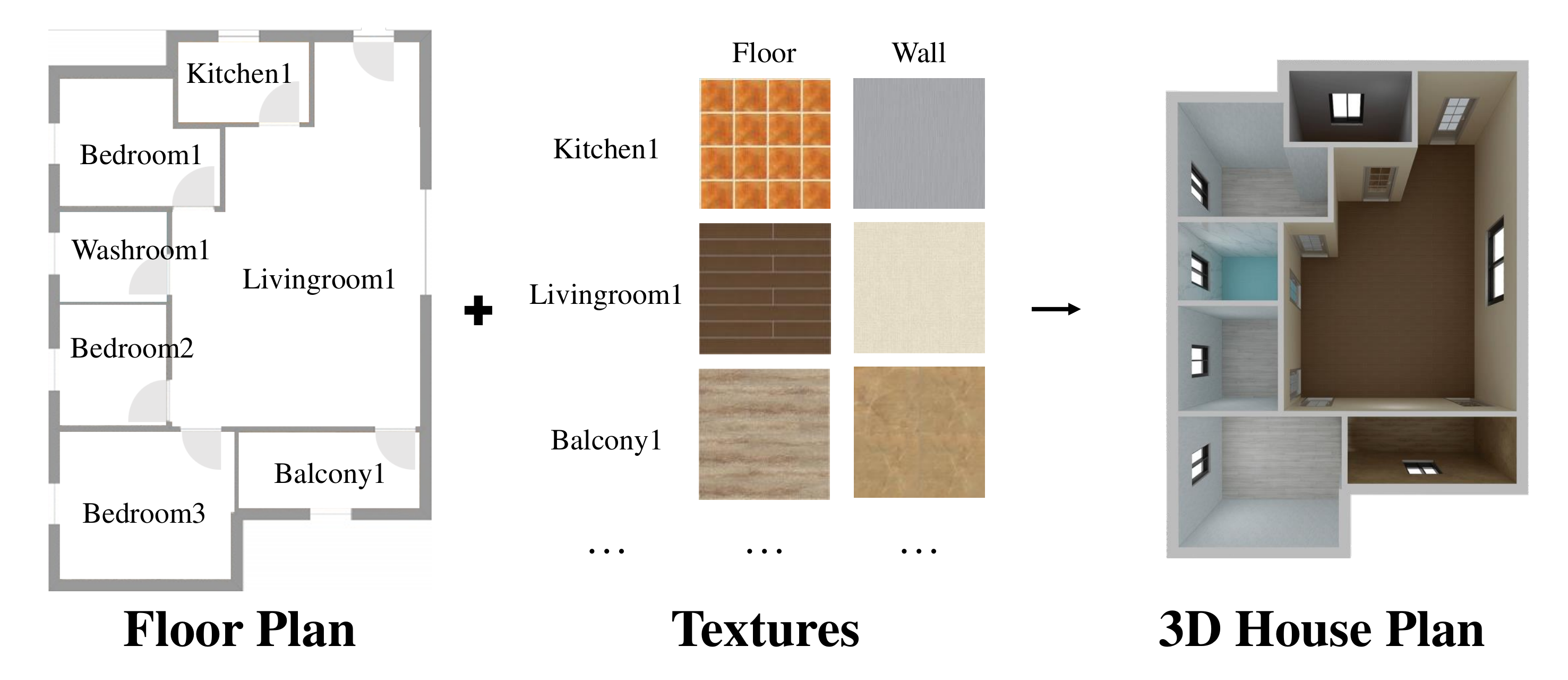}
	\end{center}
	\vspace{-20pt}
	\caption{3D house plan generation and rendering.}
	\label{fig:render_results}
	\vspace{-13pt}
\end{figure}

For the better visualisation of the generated floor plan with textures, we introduce a 3D scene generator followed with a photo-realistic rendering process.
Given generated floor plan and textures as shown in Figure~\ref{fig:render_results}, we generate walls from boundaries of rooms with fixed height and thickness. We set the height of walls to 2.85m and the thickness of interior walls to 120mm. The thickness of the exterior wall is set to 240mm while the length of the door is 900mm and the height is 2000mm. We simply set the length of the window to thirty percent of the length of the wall it belongs to.
Besides, we develop a photo-realistic rendering based on Intel Embree~\cite{wald2014embree}, an open-source collection of high-performance ray tracing kernels for x86 CPUs. Photo-realistic renderer is implemented with Monte Carlo path tracing. By following the render equation~\cite{kajiya1986rendering}, the path tracer simulates real-world effects such as realistic material appearance, soft shadows, indirect lighting, ambient occlusion and global illumination. In order to visualise the synthetic scenes, we deploy a virtual camera on the front top of each scene and capture a top-view render image. 

\section{Experiments}


\subsection{Experimental Settings} \label{experimental_settings}


\noindent\textbf{Dataset.}
To generate 3D building models from natural language descriptions, we collect a new dataset, 
which contains $2,000$ houses, $13,478$ rooms and $873$\footnote{Some rooms have same textures so this number is smaller than the total number of rooms.} texture images with corresponding natural language descriptions. These descriptions are firstly generated from some pre-defined templates and then refined by human workers. The average length of the description is $173.73$ and there are $193$ unique words.
In our experiments, we use $1,600$ pairs for training while $400$ for testing in the building layout generation. For texture synthesis, we use $503$ data for training and $370$ data for testing. More dataset analysis can be found in the supplementary materials.

\vspace{5pt}
\noindent\textbf{Evaluation metrics.}
We quantitatively evaluate our model and compare it with other models in threefold: layout generation accuracy, texture synthesis performance, and final 3D house floor plans. We measure the precision of the generated layout by \textit{Intersection-over-Union (IoU)}, 
which indicates the overlap between the generated box and ground-truth one, where the value is from 0 to 1. 
For the evaluation of textures, we use \textit{Fr\'echet Inception Distance (FID)}~\cite{heusel2017gans}. In general, the smaller this value is, the better performance the method will have.
Besides, to test the pair-wise similarity of generated images and identify mode collapses reliably~\cite{odena2017conditional}, we use \textit{Multi-scale Structural Similarity  (MS-SSIM)}~\cite{wang2003multiscale} for further validation. A lower score indicates a higher diversity of generated images (\ie, fewer model collapses).
Note that, following the settings in~\cite{Han17stackgan2}, for a fair comparison, we resize all the images to $64\times64$ before computing FID and MS-SSIM. For the 3D house floor plans, which are our final products, we run a human study to evaluate them.

%

\vspace{5pt}
\noindent\textbf{Implementation details.}
In practice, we set input $\mathbf{Z}\in\mathbb{R}^{w\times h\times (d_1+d_2+d_3)}$ of LCT-GAN with $h=5$, $w=5$, $d_1=100$, $d_2=19$ and $d_3=12$.
All the weights of models (GC-LPN and LCT-GAN) are initialised from a normal distribution with zero-mean and standard deviation of $0.02$.
In training, we use Adam~\cite{kingma2014adam} with $\beta_1=0.5$ to update the model parameters of both GC-LPN and LCT-GAN.
We optimise our LCT-GAN to generate texture images of size $160\times160$ with mini-batch size $24$ and learning rate $0.0002$.

\subsection{Building Layout Generation Results} \label{sec:build_layout_generation}

\paragraph{Compared methods.}

We evaluate the generated layout and compare the results with baseline methods. However, there is no existing work on our proposed text-guided layout generation task, which focuses on generating building layouts directly from given linguistic descriptions. Therefore, our comparisons are mainly to ablated versions of our proposed network. The compared methods are:

\textbf{MLG:} In ``Manually Layout Generation'' (MLG), we draw the building layouts directly using a program with the predefined rules, according to the given input attributes, such as type, position and size of the rooms. Specifically, we first roughly locate the central coordinates of each room conditioning on the positions. After that, we randomly pick the aspect ratio $\rho\in(\frac{2}{3}, \frac{3}{2})$ for different rooms, and then get the exact height and width by considering the size of each room. Finally, we draw the building layouts with such centre, height, width and type of each room.

\textbf{C-LPN:} In ``Conditional Layout Prediction Network'' (C-LPN), we simply remove the GCN in our proposed model. That means, when generating building layouts, the simplified model can only consider the input descriptions and ignore the information from neighbourhood nodes.

\textbf{RC-LPN:} In ``Recurrent Conditional Layout Prediction Network'' (RC-LPN), we yield the outline box of rooms sequentially like~\cite{tan2019text2scene}. To be specific, we replace GCN with an LSTM and predict the building layout by tracking the history of what has been generated so far.

\vspace{-10pt}
\paragraph{Quantitative evaluation.}

\begin{table}[t]
	\centering
	\begin{tabular}{c|c|c|c|c}
		\hline
		& MLG & C-LPN & RC-LPN & GC-LPN (ours) \\
		\hline
		IoU & 0.7208 & 0.8037 & 0.7918 & \textbf{0.8348} \\
		\hline
	\end{tabular}%
	\vspace{-5pt}
	\caption{IoU results on Text-to-3D House Model dataset.}
	\vspace{-16pt}
	\label{tab:IoU}%
\end{table}%

We evaluate the performance of our proposed GC-LPN by calculating the average IoU value of the generated building layouts. From Table~\ref{tab:IoU}, compared with the baseline methods, GC-LPN obtains higher value in IoU, which implies that the GC-LPN has the capacity to locate the outline of layout more precisely than other approaches. Models without our graph-based representation, such as C-LPN and RC-LPN, have lower performance.

\begin{figure}[t]
	\vspace{-5pt}
	\begin{center}
		\includegraphics[width=1.0\linewidth]{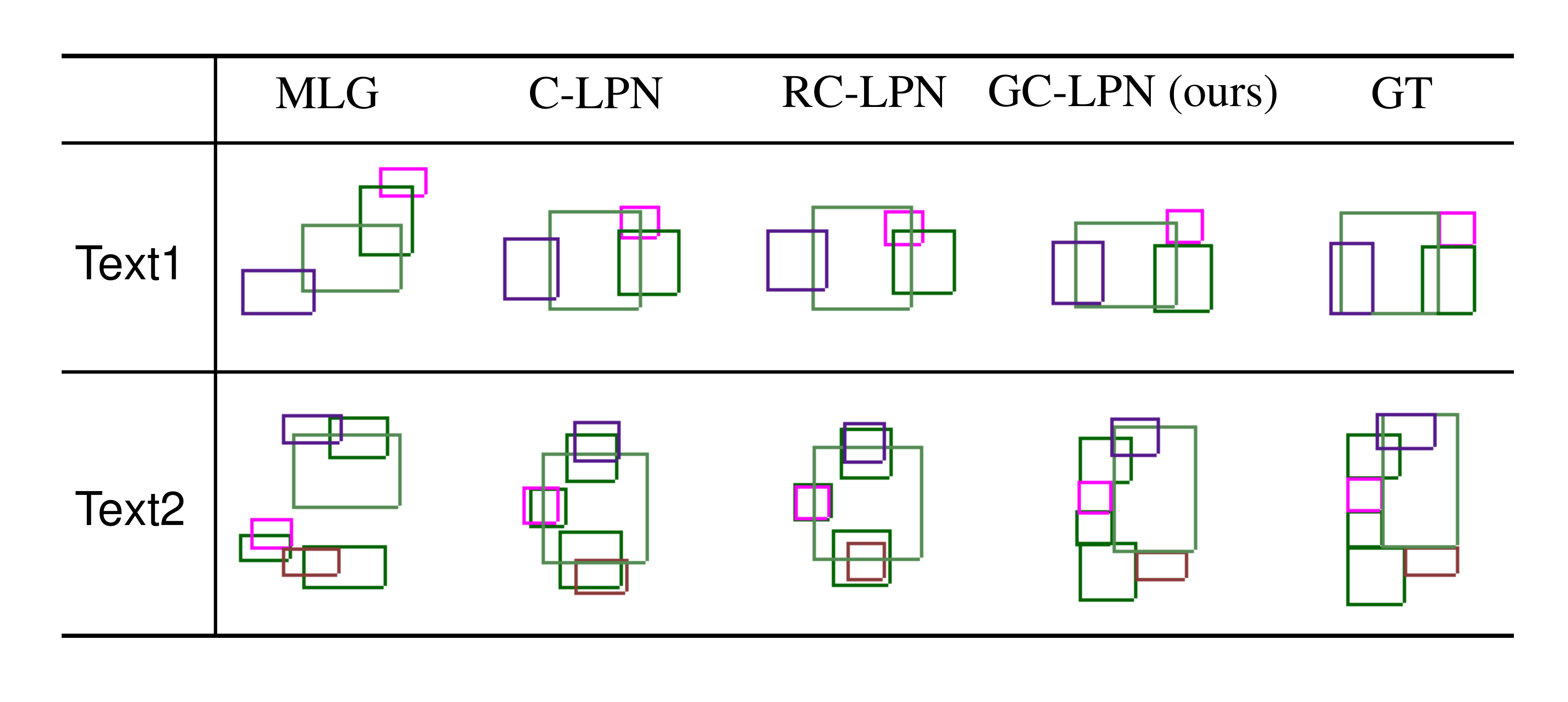}
	\end{center}
	\vspace{-25pt}
	\caption{Visual comparisons between GC-LPN and baselines. ``Text1'' and ``Text2''
		are the input descriptions, where ``Text1'' is relatively simple while ``Text2'' is more complex.
	}
	\label{fig:layout_quality_comparison}
\end{figure}

\begin{figure}[t]
	\begin{center}
		\vspace{-5pt}
		\includegraphics[width=0.80\linewidth]{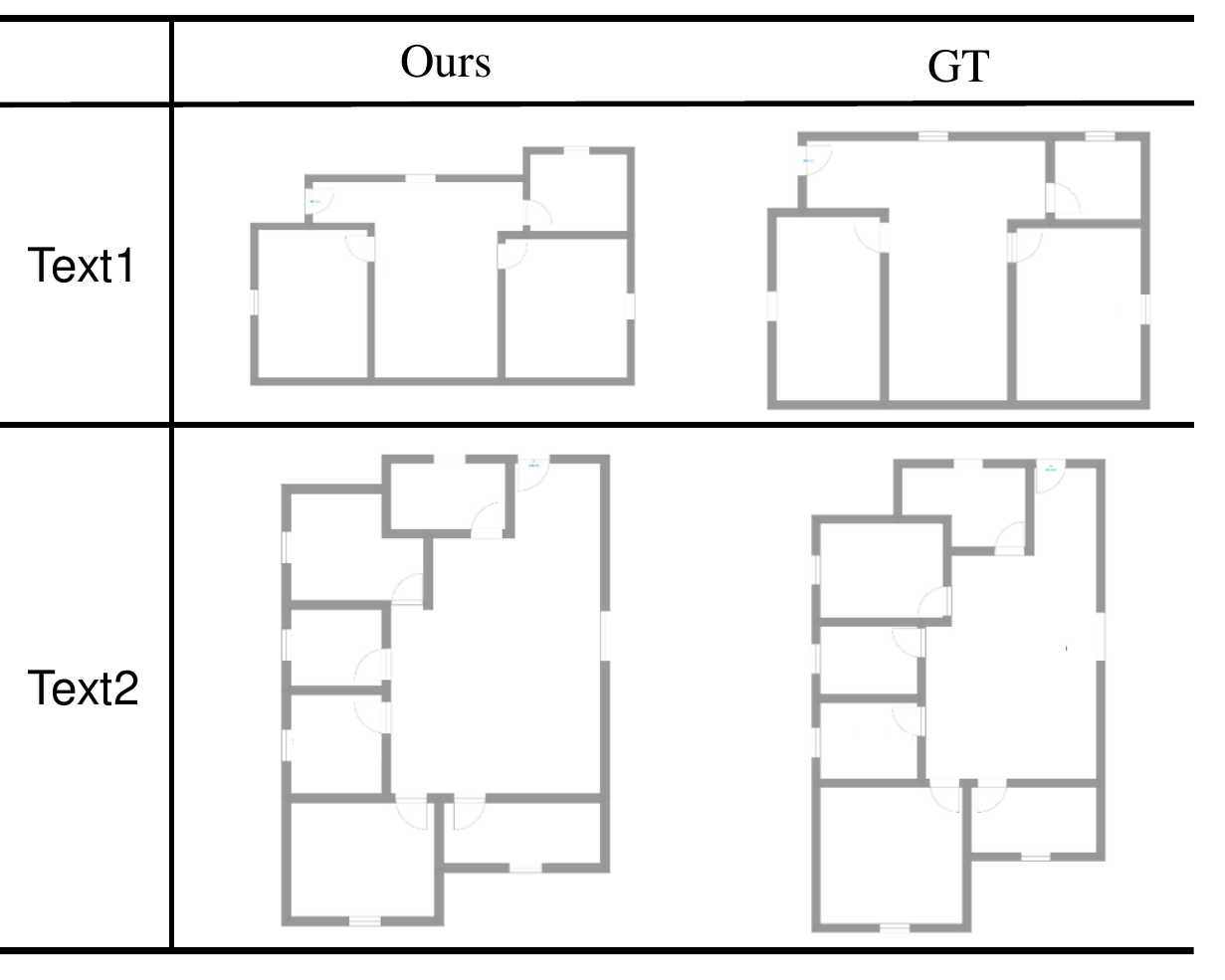}
	\end{center}
	\vspace{-20pt}
	\caption{Examples of generated 2d floor plans and ground-truth counterparts with ``Text1'' and ``Text2'', respectively.
	}
	\vspace{-12pt}
	\label{fig:example_2d_floor_plan}
\end{figure}

\vspace{5pt}
\noindent\textbf{Qualitative evaluation.}
Moreover, we investigate the performance of our GC-LPN by visual comparison. From Figure~\ref{fig:layout_quality_comparison}, we provide two layout samples corresponding to ``Text1\footnote{\emph{The building contains one washroom, one bedroom, one livingroom, and one kitchen. Specifically, washroom1 has 5 squares in northeast. bedroom1 has 14 square meters in east. Besides, livingroom1 covers 25 square meters located in center. kitchen1 has 12 squares in west. bedroom1, kitchen1, washroom1 and livingroom1 are connected. bedroom1 is next to washroom1. 
}}''
and ``Text2\footnote{Due to page limit, we put the content of ``Text2'' in the supplementary.}''
respectively. The results show that compared with the baseline methods, GC-LPN obtains more accurate layouts, whether simple or complex.
We also present the generated 2D floor plans after post-processing, and the corresponding ground-truths in Figure~\ref{fig:example_2d_floor_plan}.



\subsection{Texture Synthesis Results} \label{sec:texture_generation}

\paragraph{Compared methods.}
For the conditional texture generation task, we compare the performance of our proposed method with several baselines, including ACGAN~\cite{odena2017conditional}, StackGAN-v2~\cite{Han17stackgan2} and PSGAN~\cite{bergmann2017learning}. Note that PSGAN can only generate image from random noise. Thus, to generate images in a controlled way, we design a variant of PSGAN, which introduces the conditional information when synthesising the objective texture like~\cite{mirza2014conditional}.

\vspace{5pt}
\noindent\textbf{Quantitative evaluation.}
In this part, we compare the performance of different methods on our proposed dataset in terms of FID
and MS-SSIM. In Table~\ref{tab:texture_quality}, our LCT-GAN achieves the best performance in FID, which implies that our method is able to yield more photo-realistic images than others. Moreover, for MS-SSIM, our LCT-GAN obtains the competitive result compared with PSGAN, which is also designed specifically for texture generation. It suggests that our method has the ability to ensure the diversity of synthesised images when preserving realism.

\begin{table}[t]
	\centering
	\resizebox{1.0\linewidth}{!}
	{
		\begin{tabular}{c|cc|cc}
			\hline
			\multirow{2}[2]{*}{Methods} & \multicolumn{2}{c|}{Train Set} & \multicolumn{2}{c}{Test Set} \\
			& FID   & MS-SSIM & FID   & MS-SSIM \\
			\hline
			ACGAN~\cite{odena2017conditional} &    198.07   &   0.4584    &   220.18    &  0.4601 \\
			StackGAN-v2~\cite{Han17stackgan2} &    182.96   &   0.6356    &   188.15    &  0.6225 \\
			PSGAN~\cite{bergmann2017learning} &   195.29    &   0.4162    &   217.12    & 0.4187 \\
			LCT-GAN (ours)  &  \textbf{119.33}     &   \textbf{0.3944}    &  \textbf{145.16}  & \textbf{0.3859} \\
			\hline
		\end{tabular}%
	}
	\vspace{-5pt}
	\caption{FID and MS-SSIM results of generated textures.}
	\vspace{-5pt}
	\label{tab:texture_quality}%
\end{table}%

\begin{figure}[t]
	\vspace{-5pt}
	\begin{center}
		\includegraphics[width=1.0\linewidth]{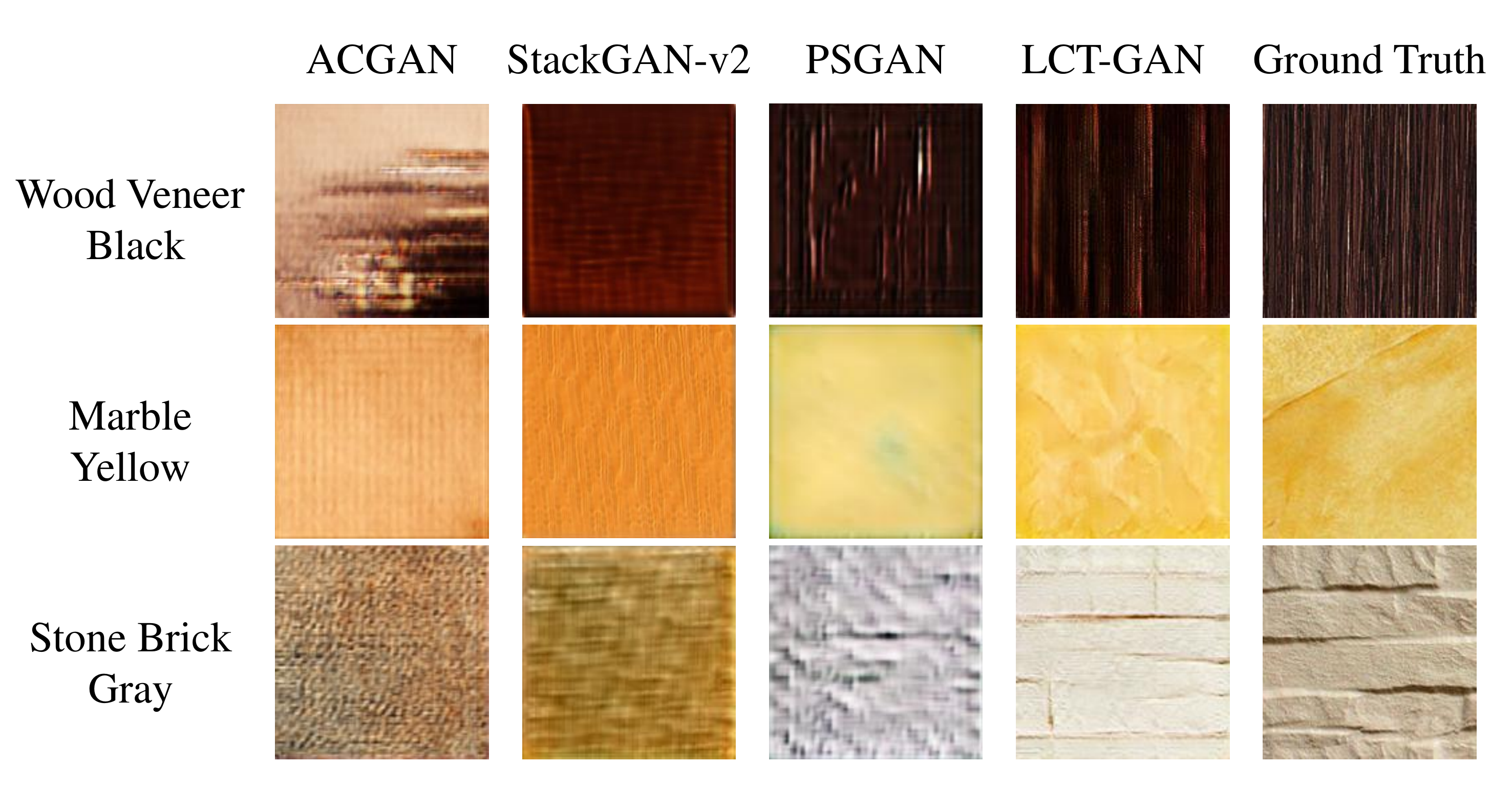}
	\end{center}
	\vspace{-20pt}
	\caption{Visual results of LCT-GAN and baselines.}
	\vspace{-12pt}
	\label{fig:texture_quality_comparison}
\end{figure}

\vspace{5pt}
\noindent\textbf{Qualitative evaluation.}
For further evaluating the performance of LCT-GAN, we provide several visual results of the generated textures. From Figure~\ref{fig:texture_quality_comparison}, compared with the baselines, our synthesised images contain more details while preserving the semantic alignment with the conditional descriptions. The results demonstrate LCT-GAN has the ability to semantically align the given texts and capture more detailed information than other approaches. 
We put more results in the supplementary.

\vspace{5pt}
\noindent\textbf{Ablation studies.}
To test the effect of each proposed loss, we conduct an ablation study to compare the generated results by removing some losses and show the quantitative results in Table~\ref{tab:texture_ablation}. 
Note that the model only using adversarial loss $\mathcal{L}_{Adv}$ can not yield controllable images. Thus, we combine $\mathcal{L}_{Adv}$ with the other two losses (\ie, $\mathcal{L}_{M}$ and $\mathcal{L}_{C}$) to investigate the performance.
The results show that based on $\mathcal{L}_{Adv}$, $\mathcal{L}_{M}$ and $\mathcal{L}_{C}$ are able to improve the performance very well. When using all the three losses into our model, we obtain the best results on both FID and MS-SSIM.

\begin{table}[t]
	\centering
	\resizebox{0.95\linewidth}{!}
	{
		\begin{tabular}{c|ccc|cc|cc}
			\hline
			\multirow{2}{*}{} & \multirow{2}{*}{$\mathcal{L}_{Adv}$} & \multirow{2}{*}{$\mathcal{L}_{M}$} & \multirow{2}{*}{$\mathcal{L}_{C}$} & \multicolumn{2}{c|}{Train Set} & \multicolumn{2}{c}{Test Set} \\
			&       &       &       & FID & MS-SSIM & FID & MS-SSIM \\
			\hline
			\multirow{3}[2]{*}{LCT-GAN}
			& $\surd$     & $\surd$     &             & 134.06 & 0.4189 & 157.01  & 0.4191\\
			& $\surd$     &       & $\surd$           & 134.61 & 0.4310 & 158.20  & 0.4263\\
			& $\surd$     &   $\surd$    &    $\surd$   &   \textbf{119.33}  & \textbf{0.3944} &   \textbf{145.16}   & \textbf{0.3859} \\
			\hline
	\end{tabular}}
	\vspace{-5pt}
	\caption{Impact of losses in conditional texture generation.}
	\vspace{-10pt}
	\label{tab:texture_ablation}%
\end{table}%

\begin{figure}[t]
	\vspace{-5pt}
	\begin{center}
		\includegraphics[width=0.90\linewidth]{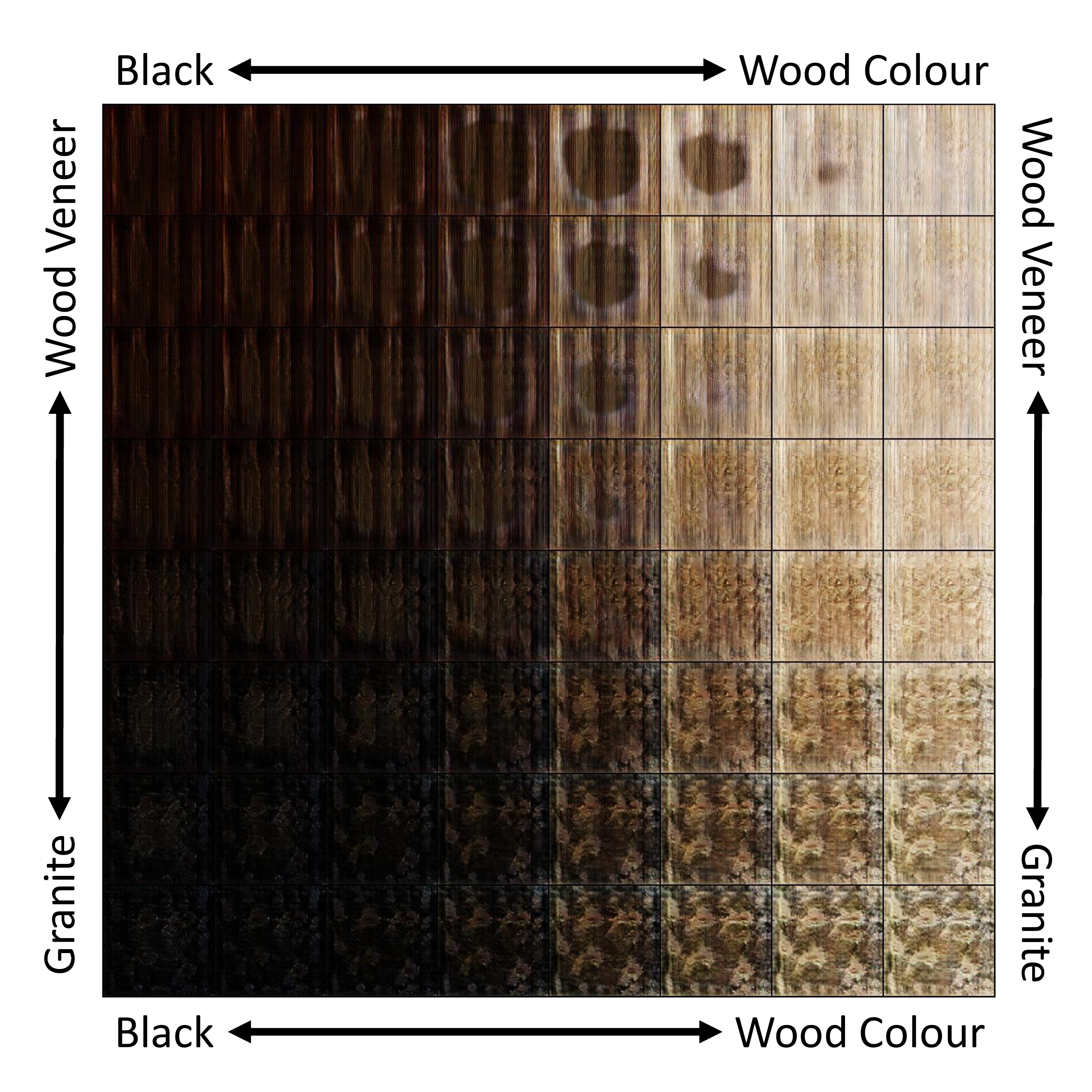}
	\end{center}
	\vspace{-23pt}
	\caption{Interpolation results of generated texture images.}
	\label{fig:interpolation}
\end{figure}

\begin{figure}[t]
	\vspace{-11pt}
	\begin{center}
		\includegraphics[width=0.85\linewidth]{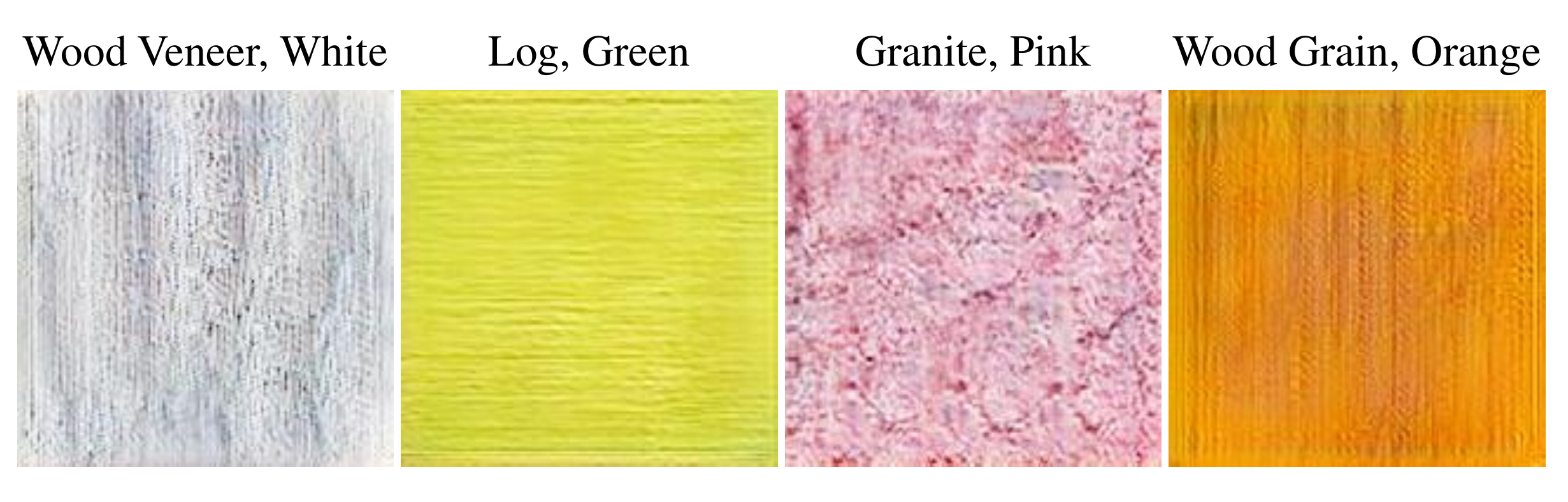}
	\end{center}
	\vspace{-20pt}
	\caption{Generated textures with novel material-colour scenarios, which are impossible existing in the real world.}
	\vspace{-10pt}
	\label{fig:novel_images}
\end{figure}

\vspace{5pt}
\noindent\textbf{Generalisation ability.}
In this section, we conduct two experiments to verify the generalisation ability of our proposed method. 
We first investigate the landscape of the latent space. Following the setting in~\cite{radford2015unsupervised}, we conduct the linear interpolations between two input embeddings and feed them into the generator $G$. As shown in Figure~\ref{fig:interpolation}, the generated textures change smoothly when the input semantics (\ie, material or colour) vary. 
On the other hand, to further evaluate the generalisation ability of our LCT-GAN, we feed some novel descriptions, which are not likely to be seen in the real world, into the generator $G$. From Figure~\ref{fig:novel_images}, even with such challenging semantic setting, our proposed method is still able to generate meaningful texture images.
Both the two experiments suggest that LCT-GAN generalises well to novel/unseen images rather than simply remembering the existing data in the training set. 


\subsection{3D House Design} \label{sec:3d_house_plan}

\begin{figure}[t]
	\begin{center}
		\includegraphics[width=0.75\linewidth]{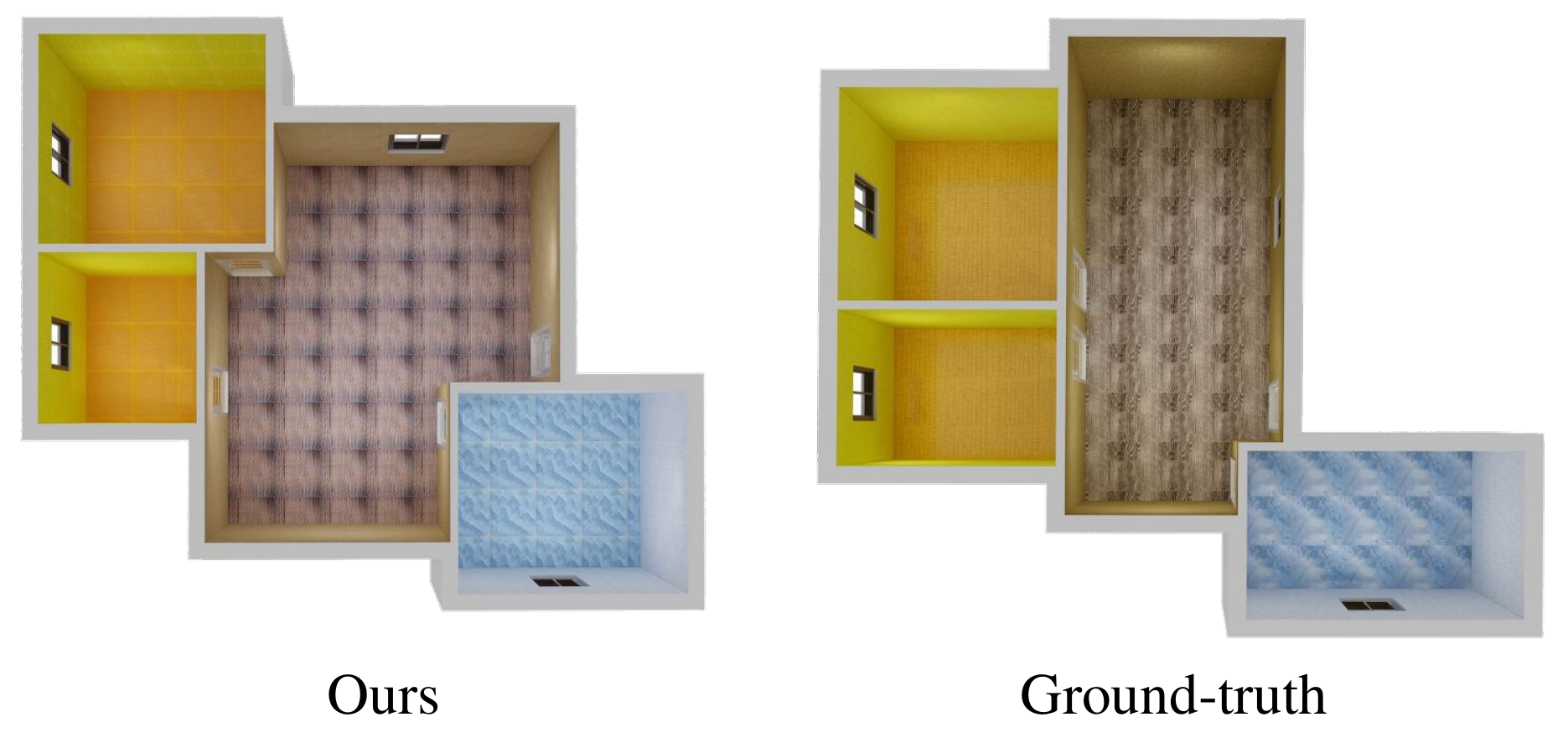}
	\end{center}
	\vspace{-18pt}
	\caption{Comparison of our generated 3D house plans with ground-truth (human-made) counterparts.}
	\vspace{-15pt}
	\label{fig:qualitative_3d_house_plan}
\end{figure}

\paragraph{Qualitative results.}
For quality evaluation, we present the 3D house plans (in Figure~\ref{fig:qualitative_3d_house_plan}) generated by our HPGM and the ground-truth counterparts with conditional text\footnote{\emph{Building layout contains one washroom, one study, one livingroom, and one bedroom. To be specific, washroom1 has Blue Marble floor, and wall is Wall\_Cloth and White. washroom1 is in southeast with 11 square meters. Additionally, study1 has Wood\_color Log floor as well as has Yellow Wall\_Cloth wall. study1 has 8 squares in west. livingroom1 is in center with 21 square meters. livingroom1 wall is Earth\_color Wall\_Cloth while uses Black Log for floor. Besides, bedroom1 covers 10 square meters located in northwest. bedroom1 floor is Wood\_color Log, and has Orange Pure\_Color\_Wood wall. livingroom1 is adjacent to washroom1, bedroom1, study1. bedroom1 is next to study1. }}, where the floor plan and corresponding room textures are drawn by architects. Our method has ability to produce competitive visual results, even compared with the human-made plans. We will put more results in the supplementary.


\begin{table}[t]
	\centering
		\resizebox{0.65\linewidth}{!}
	{
		\begin{tabular}{c|ccc}
			\hline
			& HPGM (ours) & Human & Tie \\
			\hline
			Choice (\%)  &   39.41   &  47.94  & 12.65 \\
			\hline
		\end{tabular}%
	}
	\vspace{-5pt}
	\caption{Results of HPGM v.s. human. ``Tie'' refers to the confusing results, which can not be clearly distinguished.}
	\vspace{-10pt}
	\label{tab:human_study}%
\end{table}%

\vspace{5pt}
\noindent\textbf{Human study.} \label{sec:human_study}
Since the automatic metrics can not fully evaluate the performance of our method, we perform a human study on the house plans.
Inspired by~\cite{li2019storygan, pan2017create}, we conduct a pairwise comparison between HPGM and human beings, using $100$ house plans pairs with their corresponding descriptions. 
Then, we ask $20$ human subjects (university students) to distinguish which is designed by human beings. 
Finally, we calculate the ratio of choice and obtain the final metrics.
From Table~\ref{tab:human_study}, $39.41\%$ generated samples pass the exam, which implies that compared with the manually designed samples, the machine-generated ones are exquisite enough to confuse the evaluators.

\section{Conclusion}

\vspace{-2pt}

3D house generation from linguistic descriptions is non-trivial due to the intrinsic complexity. In this paper, we propose a novel House Plan Generative Model (HPGM), dividing the generation process into two sub-task: \textit{building layout generation} and \textit{texture synthesis}. To tackle these problems, we propose two modules (\ie, GC-LPN and LCT-GAN), which focus on producing floor plan and corresponding interior textures from given descriptions.
To verify the effectiveness of our method, we conduct a series of experiments, including quantitative and qualitative evaluations, ablation study, human study, \etc. The results show that our method performs better than the competitors, which indicates the value of our approach. We believe this will be a practical application with further polish.



{
	\small
	\bibliographystyle{ieee_fullname}

}

\clearpage


\appendix

\twocolumn[
\centering
\Large{\textbf{Intelligent Home 3D: Automatic 3D-House Design \\ 
			from Linguistic Descriptions Only \\}
			(Supplementary Material)}
		\vspace{30pt}
]





\section{Details of Scene Graph Parser}

Given a series of linguistic requirements, we extract the corpus from dataset and then construct the scene graphs based on the corresponding expressions like~\cite{schuster2015generating}. 
First, we distribute the words in corpus into three categories: object $O$, relation $R$ and attribute $A$. Given an input sentence $S$, we convert this sentence to a scene graph $G=(V, E)$. $V=\{v_1, v_2, \cdots, v_n\}$ is the objects with attributes, which have been mentioned in sentence $S$. Specifically, $v_i$ consists of $o_i$ and $A_i$ (\ie, $v_i=(o_i, A_i)$), where $o_i\in O$ denotes the object of $v_i$ while $A_i\subseteq A$ is the attribute of $v_i$. 
$E\subseteq V\times R\times V$ denotes the set of relations between two objects. Each relation $e_i=(o_j, r_i, o_k)$, where $r_i\in R$.

\paragraph{Scene graph of each room.}

\begin{figure}[h]
	\renewcommand\thefigure{A}
	\begin{center}
		\includegraphics[width=1.0\linewidth]{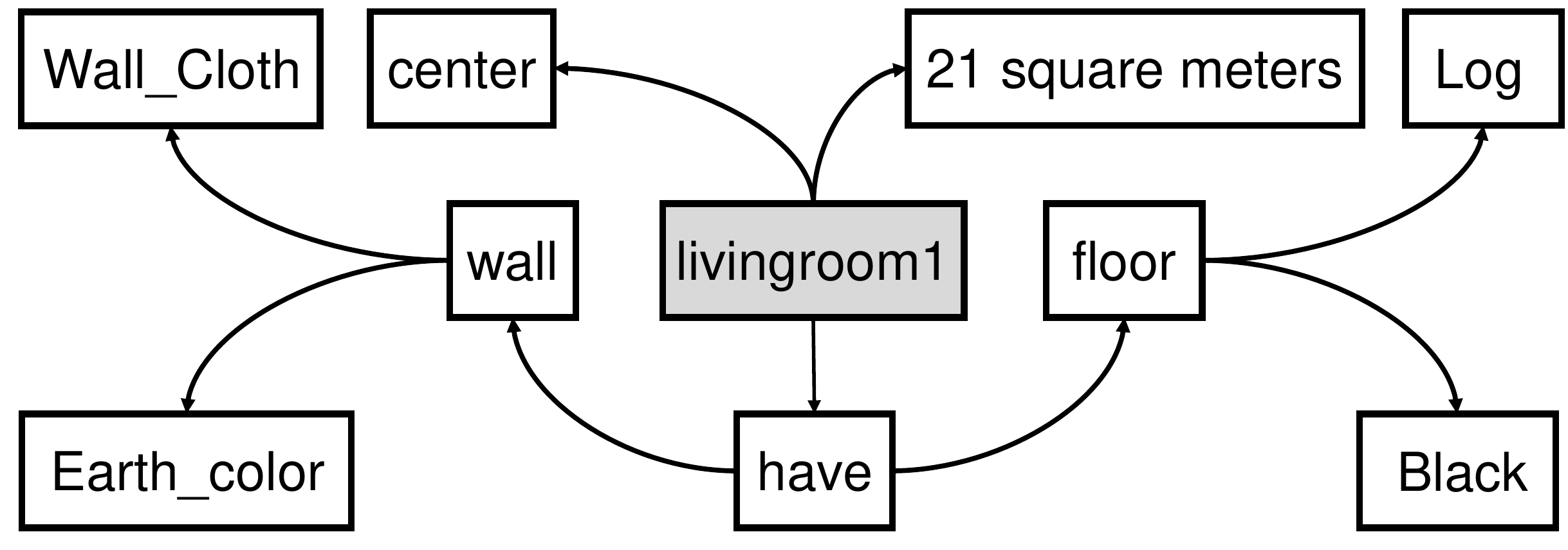}
	\end{center}
	\caption{Scene graph of \textit{``livingroom1''} according to the sentences $S_1$ and $S_2$.
	}
	\label{fig:semantic_graph}
\end{figure}

For example, for a given room (\eg, ``livingroom1''), we have the linguistic descriptions $S_1=$``\textit{livingroom1 is in center with 21 square meters}'' and $S_2=$``\textit{livingroom1 wall is Earth\_color Wall\_Cloth while uses Black Log for floor}''. We first transform $S_1$ to an object node $v_1 = (\textit{livingroom1}, \{\textit{center}, \textit{21 square meters}\})$ and the relation $e_1 = (v_1, \textit{is}, \emptyset)$.
For the sentence $S_2$, we extract the objects $v_2 = (\textit{livingroom1}, \emptyset)$, $v_3 = (\textit{wall}, \{\textit{Earth\_color}, \textit{Wall\_Cloth}\})$ and  $v_4 = (\textit{floor}, \{\textit{Black}, \textit{Log}\})$. The corresponding relations are $e_2 = (v_2, \textit{have}, v_3)$ and $e_3 = (v_2, \textit{have}, v_4)$.
Since $v_1$ and $v_2$ have the same object (\ie, \textit{``livingroom1''}), we merge them together and obtain $v_5 = (\textit{livingroom1}, \{\textit{center}, \textit{21 square meters}\})$. 
Thus, we finally get the objective scene graph (shown in Figure~\ref{fig:semantic_graph}).

\paragraph{Scene graph of adjacency between rooms.}

\begin{figure}[h]
	\renewcommand\thefigure{B}
	\begin{center}
		\includegraphics[width=1.0\linewidth]{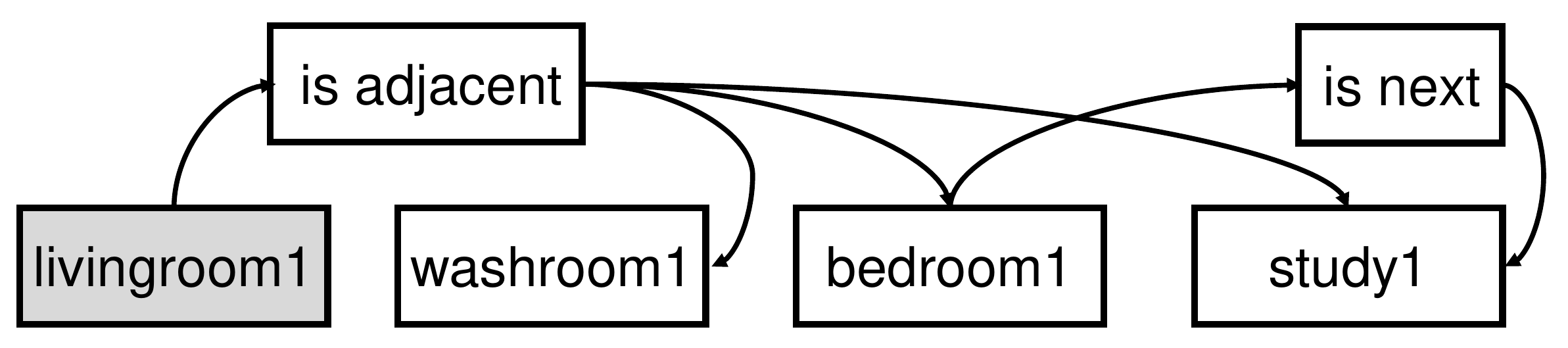}
	\end{center}
	\caption{Scene graph of adjacency between rooms according to the sentences $S_3$ and $S_4$.
	}
	\label{fig:scene_graph}
\end{figure}

In addition, the descriptions on our dataset also contain the adjacent information between the rooms, such as \textit{``livingroom1 is adjacent to washroom1, bedroom1, study1''} or \textit{``bedroom1 is next to study1''}. In order to make use of these messages, we construct another scene graph, which focuses on the relations among the rooms mentioned in given sentence. 
For example, given the sentences $S_3=$ \textit{``livingroom1 is adjacent to washroom1, bedroom1, study1''} and $S_4=$ \textit{``bedroom1 is next to study1''}, we first transform $S_3$ according to the aforementioned rules and obtain the objects and relations:
$v_{10} = (\textit{livingroom1}, \emptyset)$, $v_{11} = (\textit{washroom1}, \emptyset)$, $v_{12} = (\textit{bedroom1}, \emptyset)$ and $v_{13} = (\textit{study1}, \emptyset)$;
$e_{10} = (v_{10}, \textit{is adjacent}, v_{11})$, $e_{11} = (v_{10}, \textit{is adjacent}, v_{12})$ and $e_{12} = (v_{10}, \textit{is adjacent}, v_{13})$.
For sentence $S_4$, the objects are $v_{14} = (\textit{bedroom1}, \emptyset)$ and $v_{15} = (\textit{study1}, \emptyset)$ while the relation is $e_{13} = (v_{14}, \textit{is next}, v_{15})$. Due to $v_{12}=v_{14}$ and $v_{13}=v_{15}$, we replace $v_{14}$ and $v_{15}$ by $v_{12}$ and $v_{13}$, respectively. Therefore, $e_{13}$ can be reformulated as $e_{13} = (v_{12}, \textit{is next}, v_{13})$. We exhibit the visualised scene graph of these expressions in Figure~\ref{fig:scene_graph}.

\section{Dataset Analysis}

\begin{figure}[t]
	\renewcommand\thefigure{C}
	\begin{center}
		\includegraphics[width=1.0\linewidth]{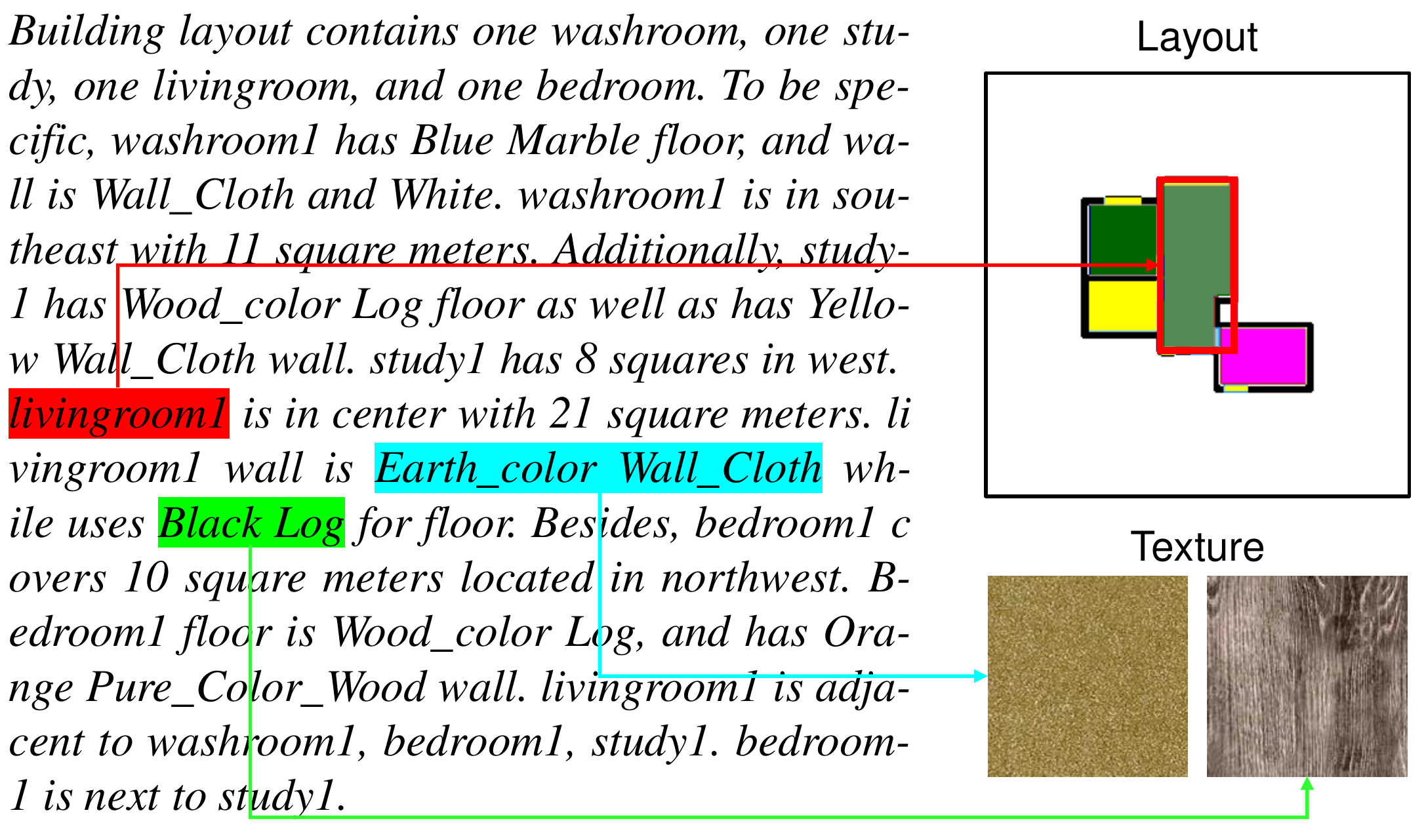}
	\end{center}
	\caption{An example from Text--to--3D House Model dataset. The given sentence contains the descriptions of both building layout and textures. For example, \textit{``livingroom1''} involves the messages about position (\textit{``center''} and size (\textit{``21 square meters''}) in linguistic expressions while a coarse bounding box covering the outline of room. The wall and floor have their descriptions (\ie, \textit{``Earth\_color Wall\_Cloth''} and \textit{``Black Log''}) and the corresponding ground-truth texture images.
	}
	\label{fig:dataset}
\end{figure}

\begin{figure}[t]
	\renewcommand\thefigure{D}
	\begin{center}
		\includegraphics[width=0.95\linewidth]{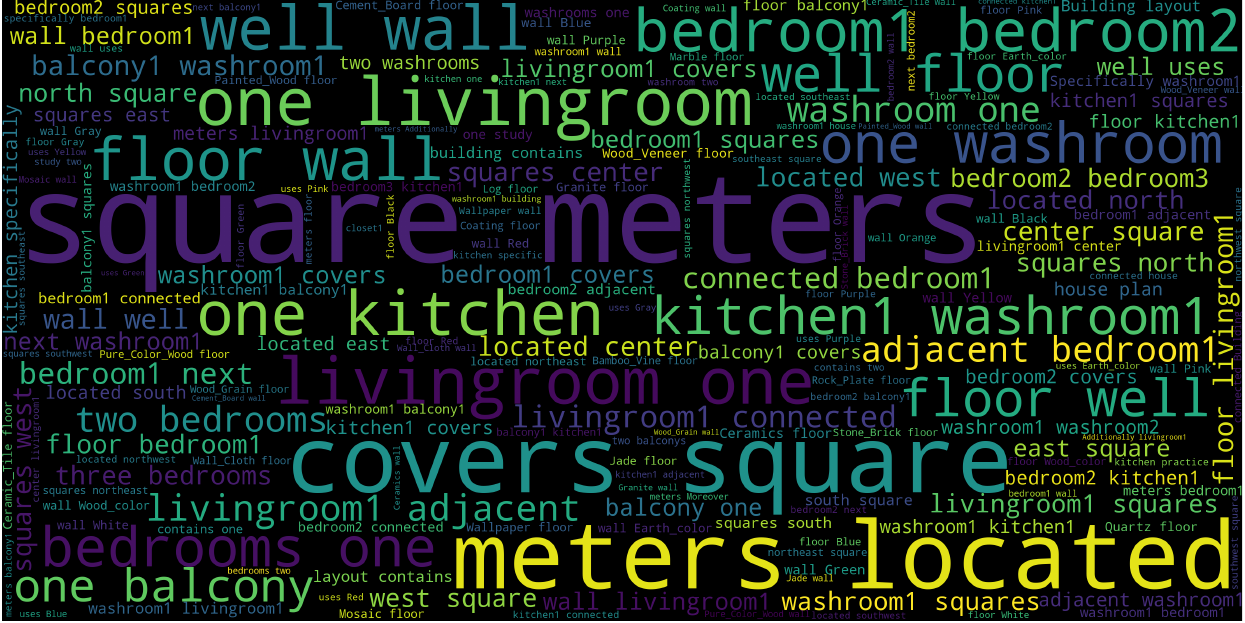}
	\end{center}
	\caption{The word cloud of the texts in our dataset.
	}
	\label{fig:word_cloud}
\end{figure}

To generate 3D building models from natural language descriptions, we collect a new Text--to--3D House Model dataset,
which contains $2,000$ houses, $13,478$ rooms and $873$\footnote{Some rooms have same textures so this number is smaller than the total number of rooms.} texture images with corresponding natural language descriptions.
These descriptions are firstly generated from some pre-defined templates and then refined by human workers. The average length of the description is $173.73$ and there are $193$ unique words.
All the building layouts are designed on the canvas with the pixel size of $512\times512$, which represents $18\times18$ square meters in the real world.
We take an example from our proposed dataset and show in Figure~\ref{fig:dataset}.
Moreover, we also provide the word cloud of our dataset and the visualised results are shown in Figure~\ref{fig:word_cloud}.

\section{Details of Generator $G$}

In this section, we provide more details of the generator $G$ in our proposed LCT-GAN and show the detailed architecture in Table~\ref{tab:model_details}.

\begin{table*}[t]
	\renewcommand\thetable{A}
	\centering
	\caption{Detailed model design of the generator $G$ of our LCT-GAN. ``F'' refers to the basic dimension of the intermediate features. ``h'' and ``w'' denote the height and width of the input, respectively.}
	{
		\begin{tabular}{c|c|c|c}
			\toprule
			Module &  Module details & Input shape    & Output shape \\
			\hline
			Upsample  &  $2\times$ Upsampling  & ($\mathrm{d_1+d_2+d_3}$, h, w) & ($\mathrm{d_1+d_2+d_3}$, 2h, 2w) \\
			\hline
			Conv2d  & kernel=(5, 5), stride=(1, 1), padding=(2, 2) & ($\mathrm{d_1+d_2+d_3}$, 2h, 2w) & (8F, 2h, 2w) \\
			\hline
			BN+ReLU & -- & (8F, 2h, 2w)  & (8F, 2h, 2w) \\
			\hline
			Upsample  &  $2\times$ Upsampling  & (8F, 2h, 2w) & (8F, 4h, 4w) \\
			\hline
			Conv2d  & kernel=(5, 5), stride=(1, 1), padding=(2, 2) & (8F, 4h, 4w) & (4F, 4h, 4w) \\
			\hline
			BN+ReLU & -- & (4F, 4h, 4w)  & (4F, 4h, 4w) \\
			\hline
			Upsample  &  $2\times$ Upsampling  & (4F, 4h, 4w) & (4F, 8h, 8w) \\
			\hline
			Conv2d  & kernel=(5, 5), stride=(1, 1), padding=(2, 2) & (4F, 8h, 8w) & (2F, 8h, 8w) \\
			\hline
			BN+ReLU & -- & (2F, 8h, 8w)  & (2F, 8h, 8w) \\
			\hline
			Upsample  &  $2\times$ Upsampling  & (2F, 8h, 8w) & (2F, 16h, 16w) \\
			\hline
			Conv2d  & kernel=(5, 5), stride=(1, 1), padding=(2, 2) &  (2F, 16h, 16w) &  (F, 16h, 16w) \\
			\hline
			BN+ReLU & -- & (F, 16h, 16w)  & (F, 16h, 16w) \\
			\hline
			Upsample  &  $2\times$ Upsampling  & (F, 16h, 16w) & (F, 32h, 32w) \\
			\hline
			Conv2d  & kernel=(5, 5), stride=(1, 1), padding=(2, 2) &  (F, 32h, 32w) &  (3, 32h, 32w) \\
			\hline
			Tanh & -- & (3, 32h, 32w)  & (3, 32h, 32w) \\
			\bottomrule
		\end{tabular}%
	}
	\label{tab:model_details}%
\end{table*}%

\section{Contents of ``Text1'' and ``Text2''}

\begin{figure}[h]
	\renewcommand\thefigure{E}
	\begin{center}
		\includegraphics[width=1.0\linewidth]{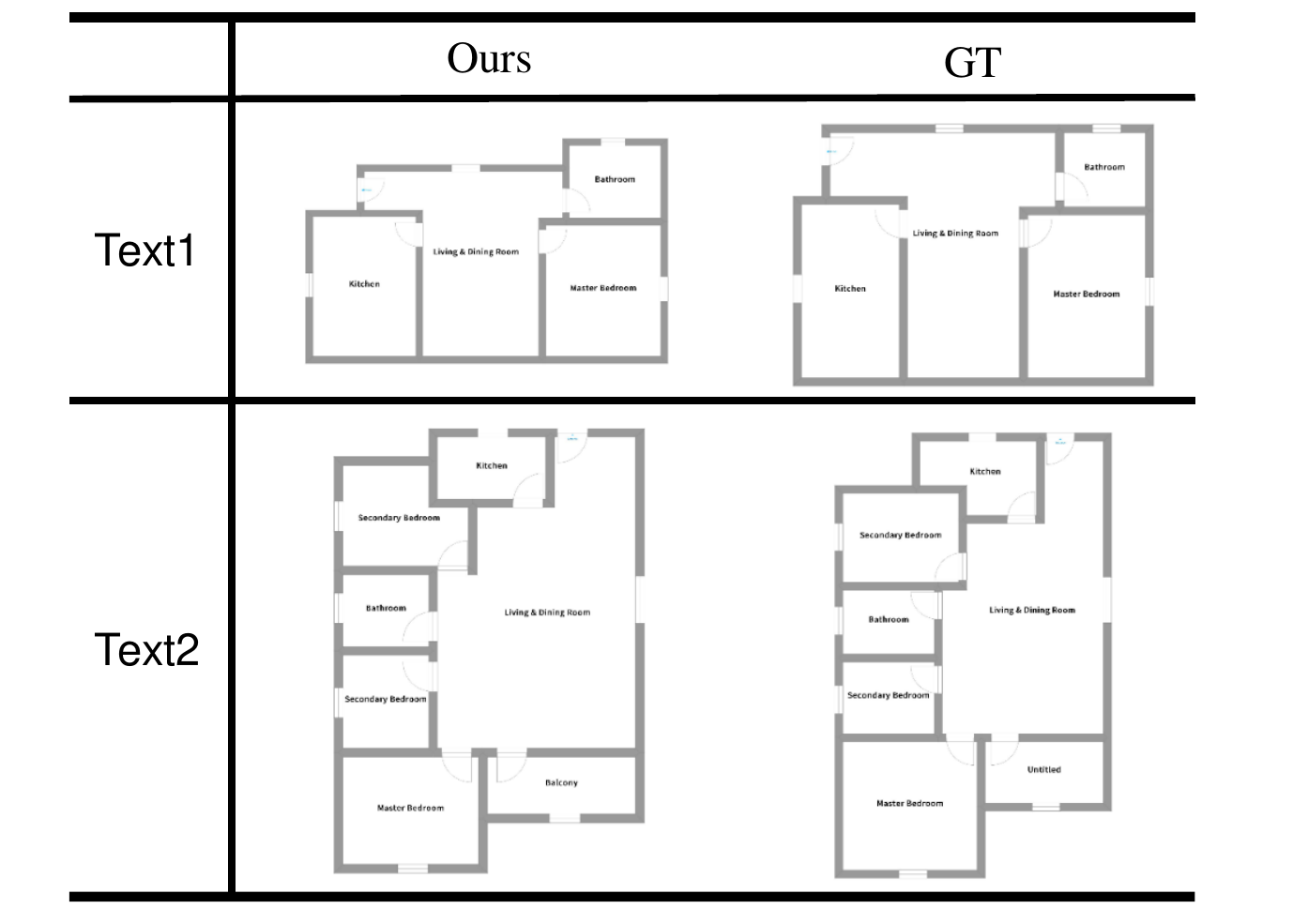}
	\end{center}
	\caption{Qualitative results of GC-LPN and ground-truth.
	}
	\label{fig:layout_quality_supp}
\end{figure}

In this section, we give the specific descriptions of ``Text1'' and ``Text2'', which have been mentioned in Figures 6 and 7 of the submitted manuscript. For convenience, in Figure~\ref{fig:layout_quality_supp}, we exhibit the visual results of 2D floor plan corresponding to ``Text1'' and ``Text2'', respectively.

\textbf{``Text1'':}
\textit{The building contains one washroom, one bedroom, one livingroom, and one kitchen. Specifically, washroom1 has 5 squares in northeast. bedroom1 has 14 square meters in east. Besides, livingroom1 covers 25 square meters located in center. kitchen1 has 12 squares in west. bedroom1, kitchen1, washroom1 and livingroom1 are connected. bedroom1 is next to washroom1.}

\textbf{``Text2'':}
\textit{The house has three bedrooms, one washroom, one balcony, one livingroom, and one kitchen. In practice, bedroom1 has 13 squares in south. bedroom2 has 9 squares in north. bedroom3 covers 5 square meters located in west. washroom1 has 4 squares in west. balcony1 is in south with 6 square meters. livingroom1 covers 30 square meters located in center. kitchen1 is in north with 6 square meters. livingroom1 is adjacent to bedroom1, bedroom2, balcony1, kitchen1, bedroom3, washroom1. balcony1, bedroom3 and bedroom1 are connected. bedroom2 is next to kitchen1, washroom1. bedroom3 is adjacent to washroom1.}

\section{More Qualitative Results}

\begin{figure*}[h]
	\renewcommand\thefigure{F}
	\begin{center}
		\includegraphics[width=1.0\linewidth]{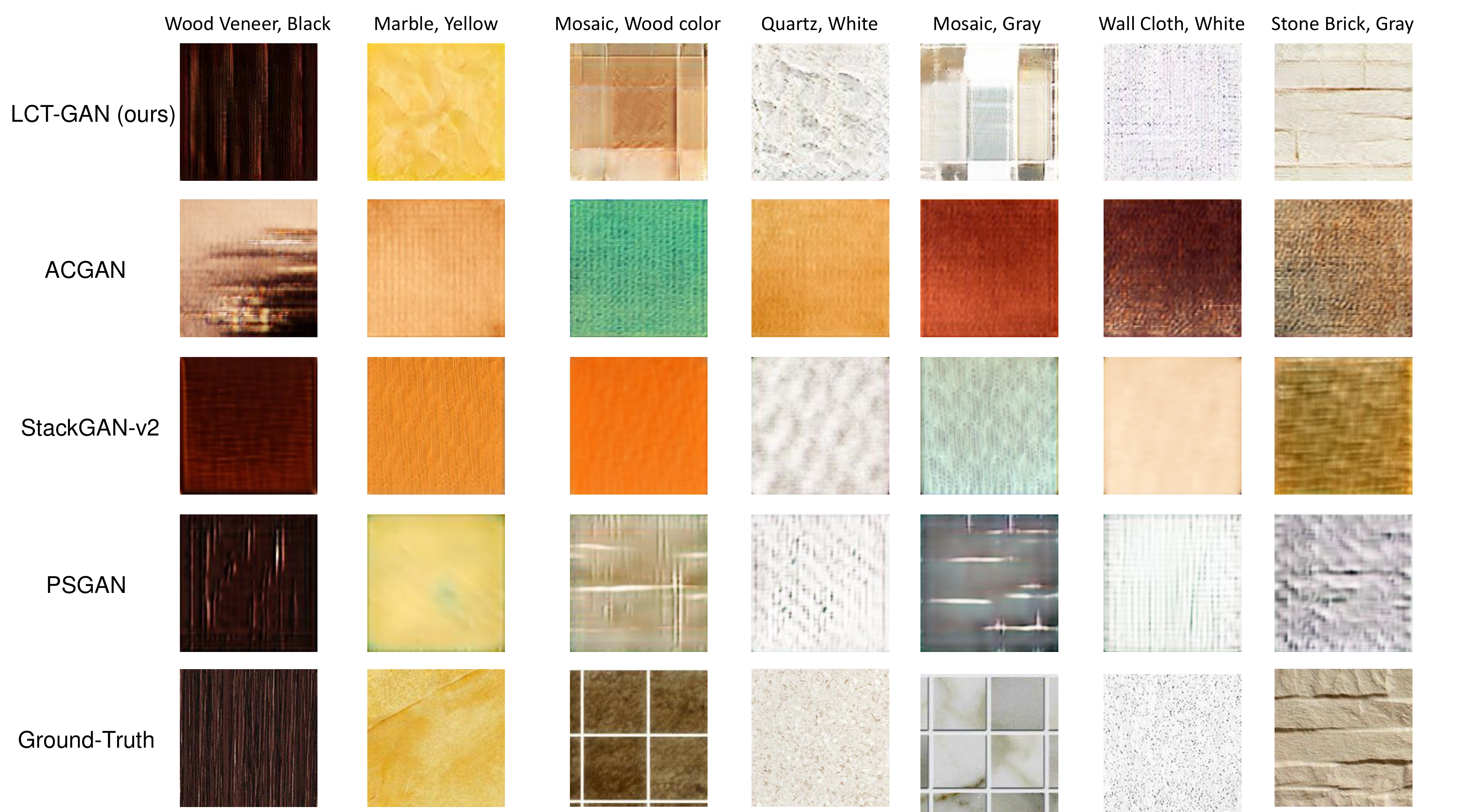}
	\end{center}
	\caption{More qualitative results of our proposed LCT-GAN and the mentioned baseline methods.
	}
	\label{fig:more_comparison_lctgan}
\end{figure*}

In this section, we will provide more qualitative results of our proposed LCT-GAN and baseline methods, which have been mentioned in the paper. 
From Figure~\ref{fig:more_comparison_lctgan}, the results show that our method is able to produce neater and sharper textures than the baseline methods. 
Besides, the generated images are more consistent with the given semantic expressions that other baselines.

\section{More Qualitative Results of 3D House Plan}

\begin{figure*}
	\renewcommand\thefigure{G}
	\centering
	{
		\scriptsize
		\resizebox{1.0\linewidth}{!}
		{
			\begin{tabular}{|>{\arraybackslash} m{4.9cm}| >{\centering\arraybackslash} m{2.8cm}| >{\centering\arraybackslash} m{2.8cm}|
					>{\centering\arraybackslash} m{2.8cm}|
					>{\centering\arraybackslash} m{2.8cm}|}
				\hline
				\centering\multirow{2}{*}{\small\textbf{Linguistic Requirements}} & \multicolumn{2}{c|}{\small\textbf{Ours}} & \multicolumn{2}{c|}{\small\textbf{Ground-truth}} \\
				\cline{2-5}
				& \small\textbf{2D Floor Plan} & \small\textbf{3D House Plan} & \small\textbf{2D Floor Plan} & \small\textbf{3D House Plan} \\
				\hline
				\textit{The building layout contains one washroom, one study, one livingroom, and one bedroom. To be specific, washroom1 has Blue Marble floor, and wall is Wall\_Cloth and White. washroom1 is in southeast with 11 square meters. Additionally, study1 has Wood\_color Log floor as well as has Yellow Wall\_Cloth wall. study1 has 8 squares in west. livingroom1 is in center with 21 square meters. livingroom1 wall is Earth\_color Wall\_Cloth while uses Black Log for floor. Besides, bedroom1 covers 10 square meters located in northwest. bedroom1 floor is Wood\_color Log, and has Orange Pure\_Color\_Wood wall. livingroom1 is adjacent to washroom1, bedroom1, study1. bedroom1 is next to study1.} 
				&\includegraphics[width=0.35\columnwidth]{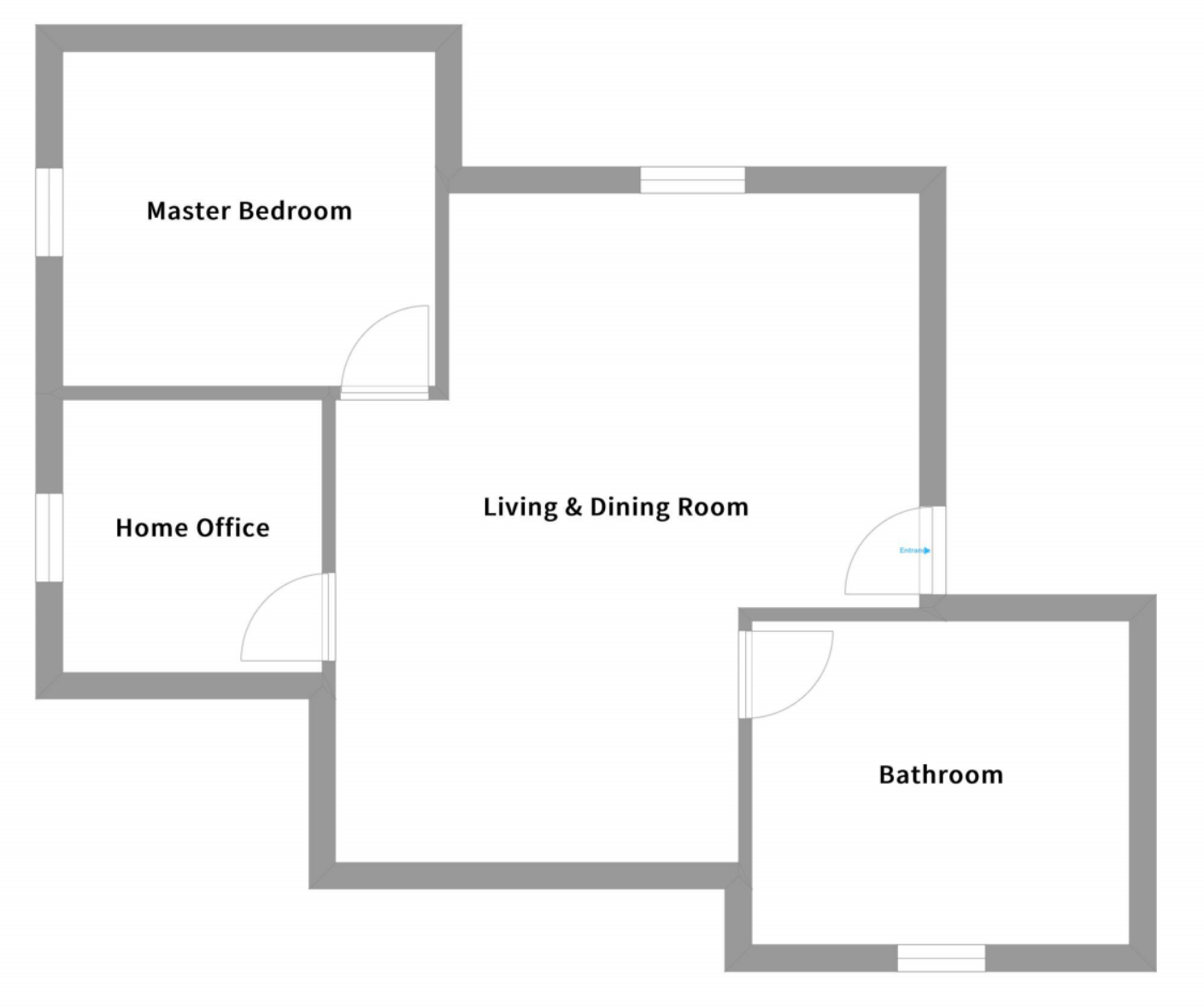} 
				&\includegraphics[width=0.35\columnwidth]{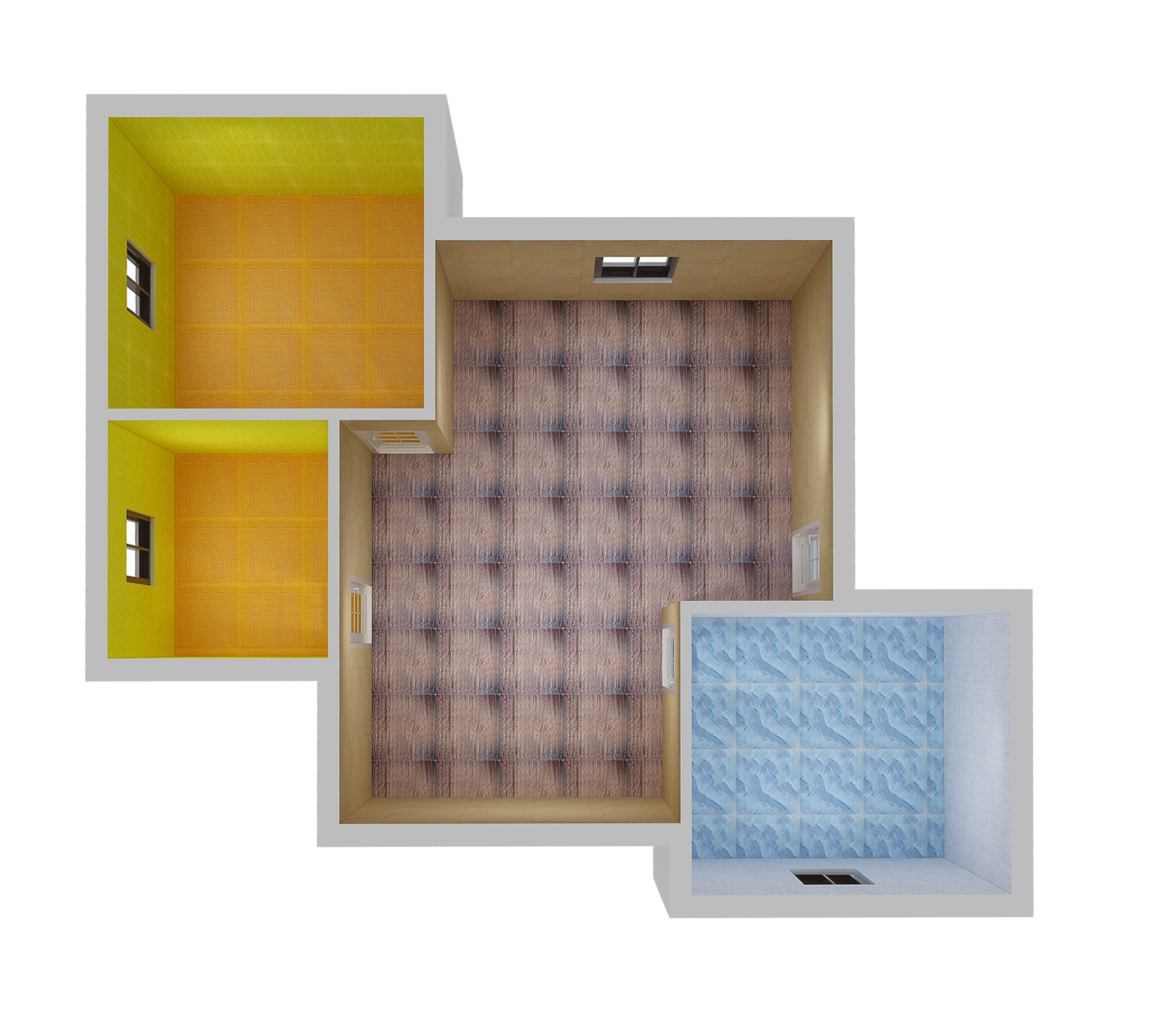}
				&\includegraphics[width=0.35\columnwidth]{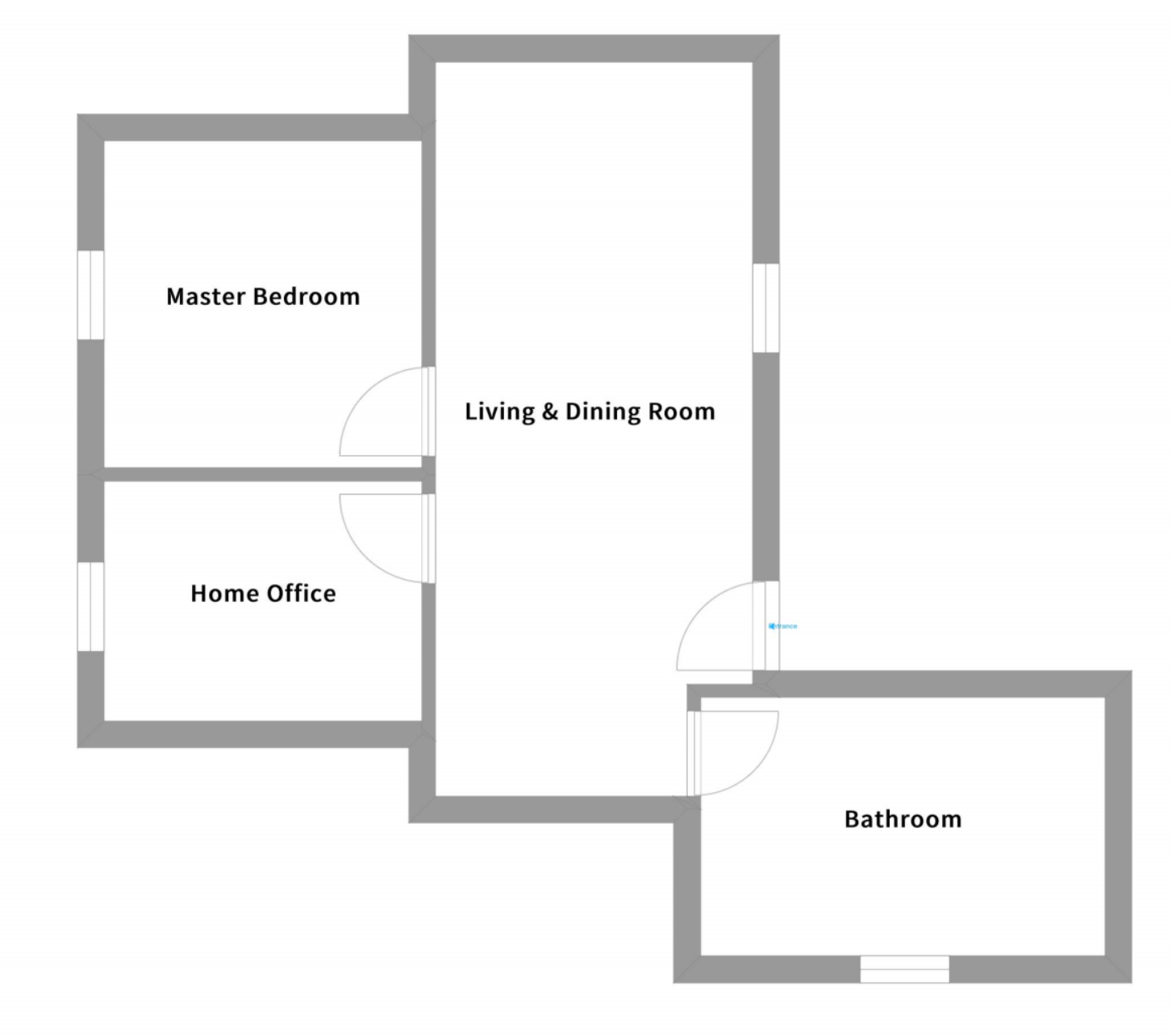}
				&\includegraphics[width=0.35\columnwidth]{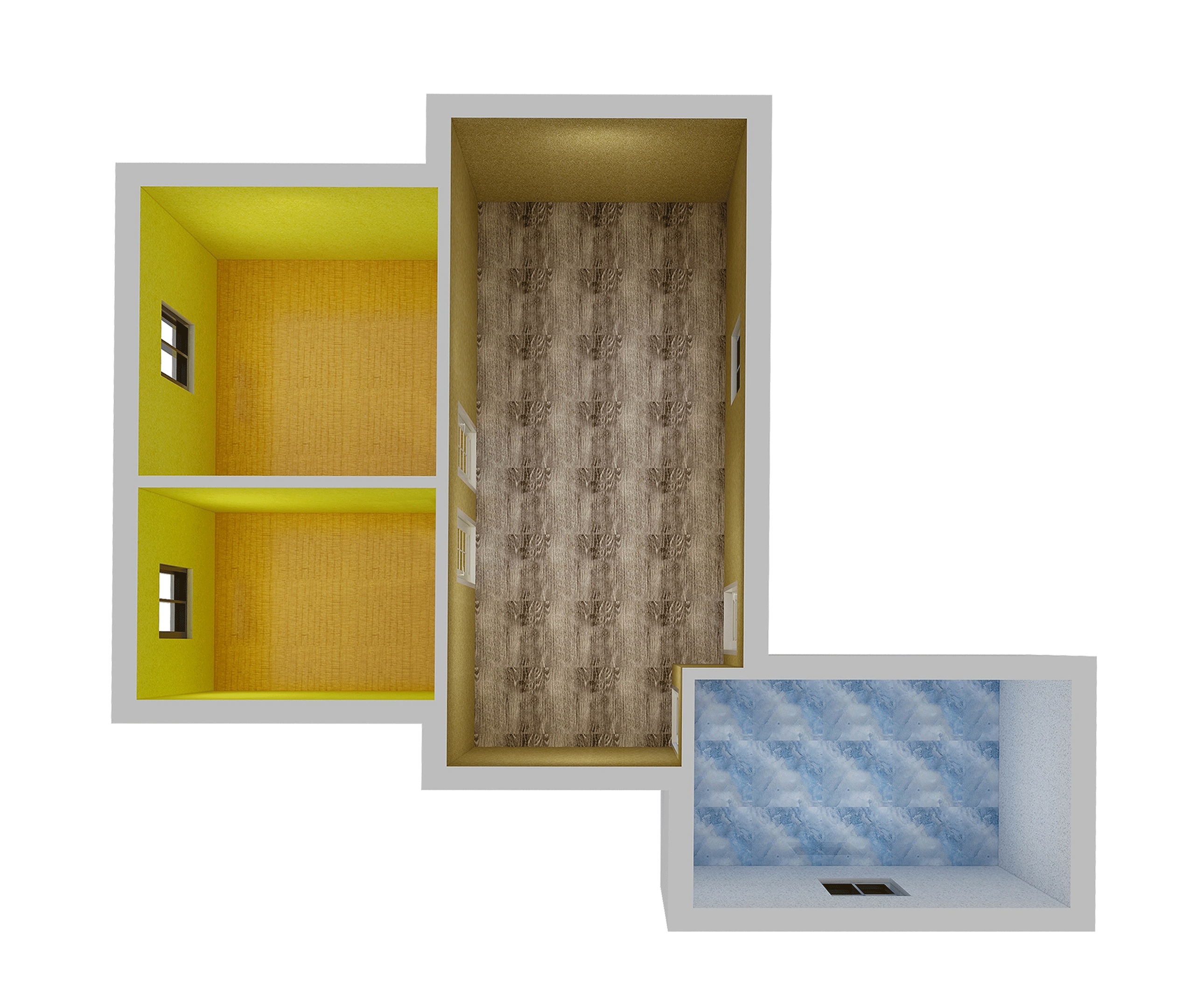} \\
				\hline
				\textit{The house has one washroom, one livingroom, one storage, two bedrooms, one kitchen, and one balcony. In practice, washroom1 is in west with 5 square meters. wall of washroom1 is Coating and Yellow, and has White Wood\_Veneer floor. Moreover, livingroom1 has Yellow Marble floor as well as wall is Wall\_Cloth and Black. livingroom1 is in center with 26 square meters. Besides, storage1 is in northwest with 9 square meters. storage1 has Wood\_color Wood\_Grain floor while wall is White Wall\_Cloth. bedroom1 has 13 squares in southwest. bedroom1 uses Black Log for floor, and wall is White Wall\_Cloth. bedroom2 uses Black Log for floor while has White Wall\_Cloth wall. bedroom2 has 7 squares in northeast. Moreover, kitchen1 uses White Wood Veneer for floor, and wall is Coating and Yellow. kitchen1 has 5 squares in north. balcony1 has 5 squares in south. balcony1 has Black Wall\_Cloth wall as well as has Yellow Marble floor. livingroom1 is adjacent to bedroom1, storage1, bedroom2, washroom1, balcony1, kitchen1. washroom1, balcony1 and bedroom1 are connected. storage1 is next to washroom1, kitchen1. bedroom2 is adjacent to kitchen1.} 
				&\includegraphics[width=0.35\columnwidth]{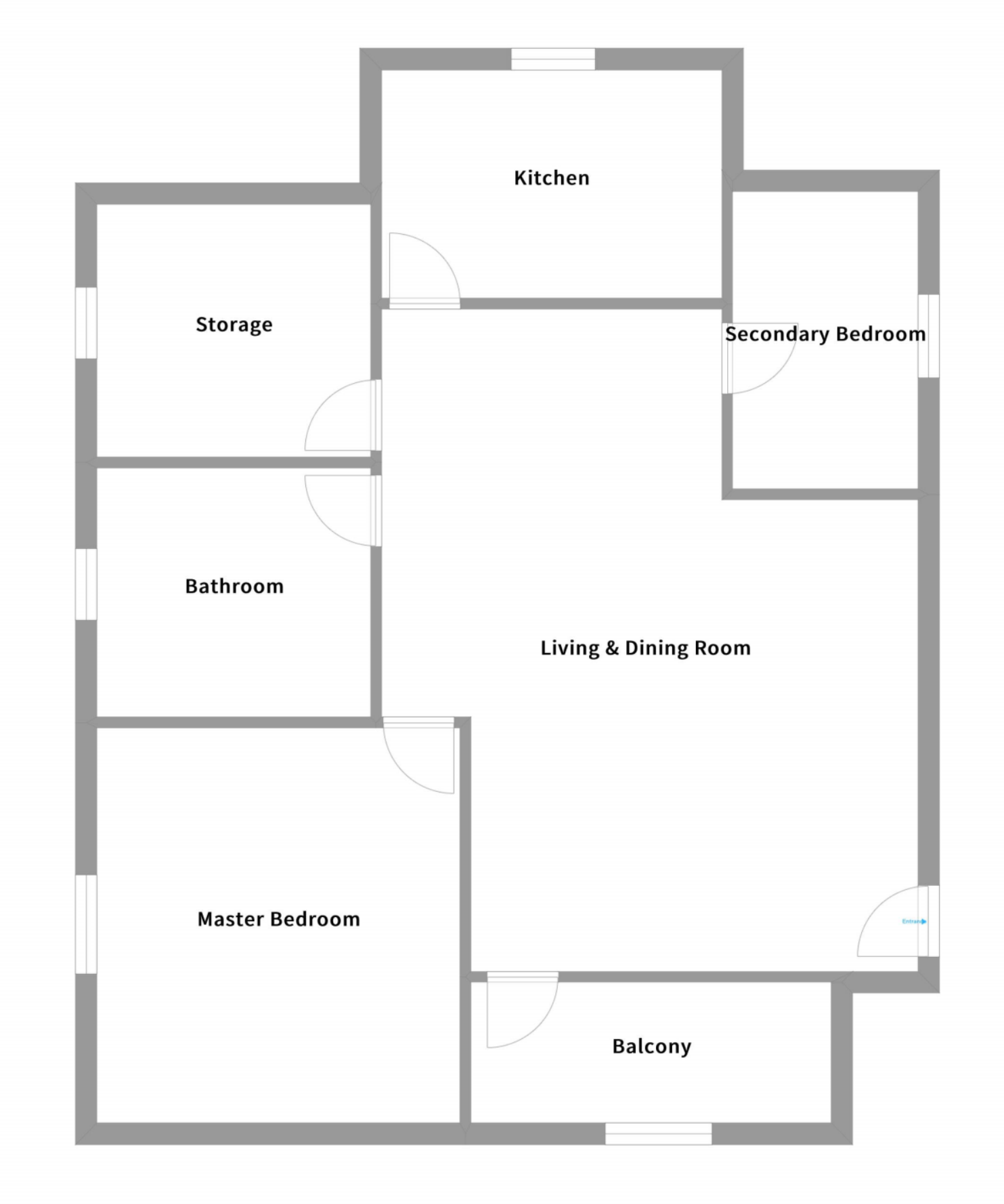} 
				&\includegraphics[width=0.35\columnwidth]{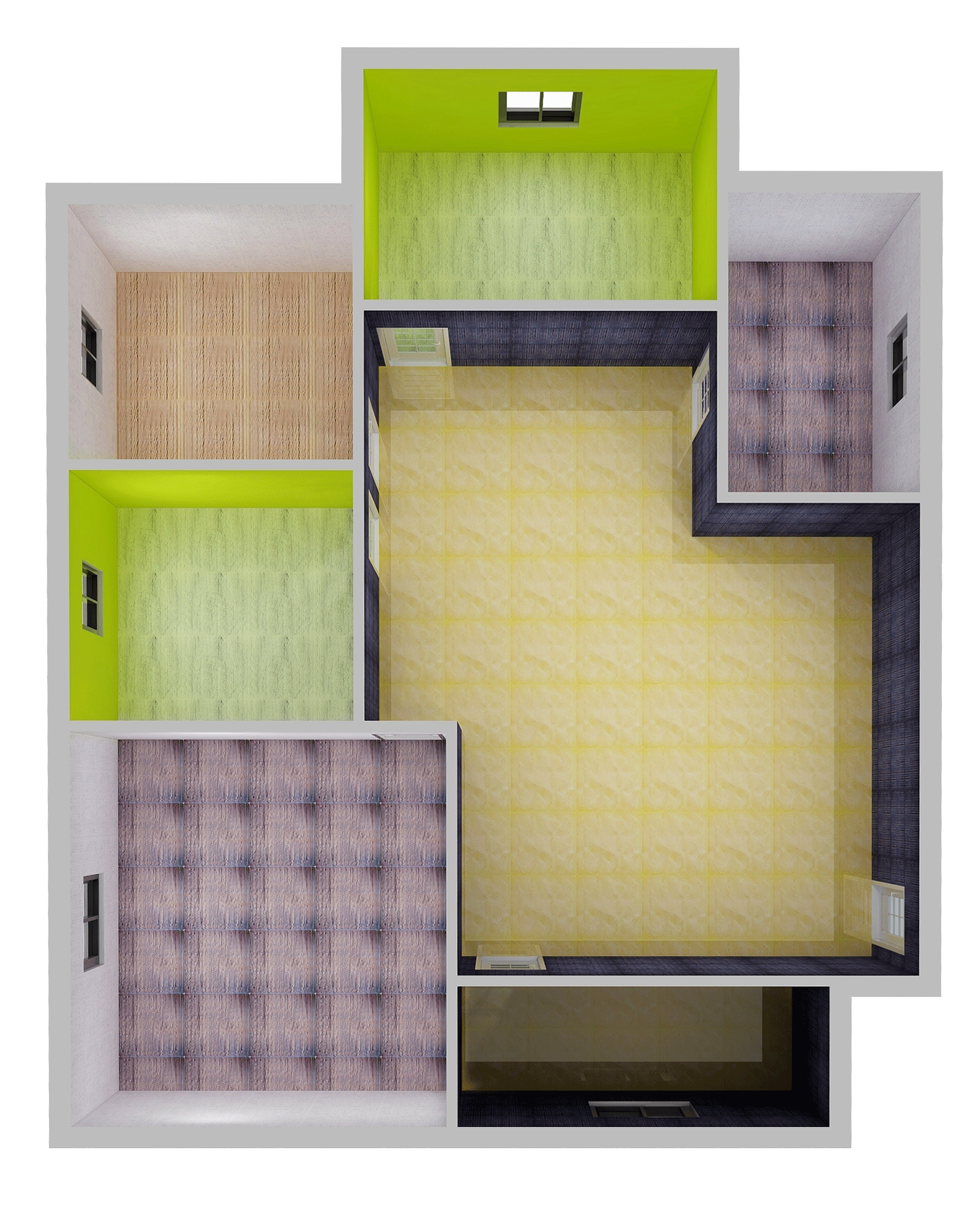}
				&\includegraphics[width=0.35\columnwidth]{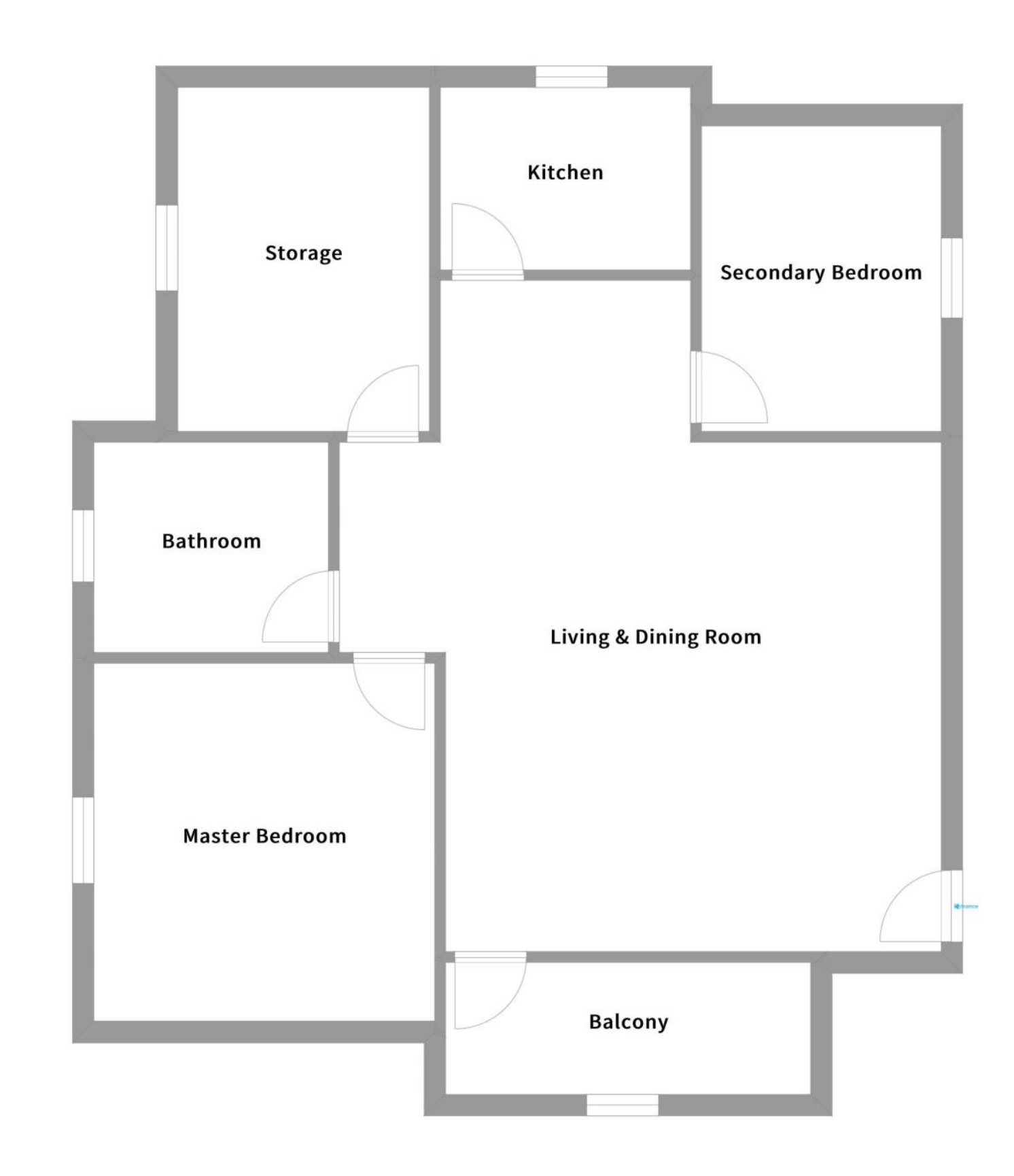} 
				&\includegraphics[width=0.35\columnwidth]{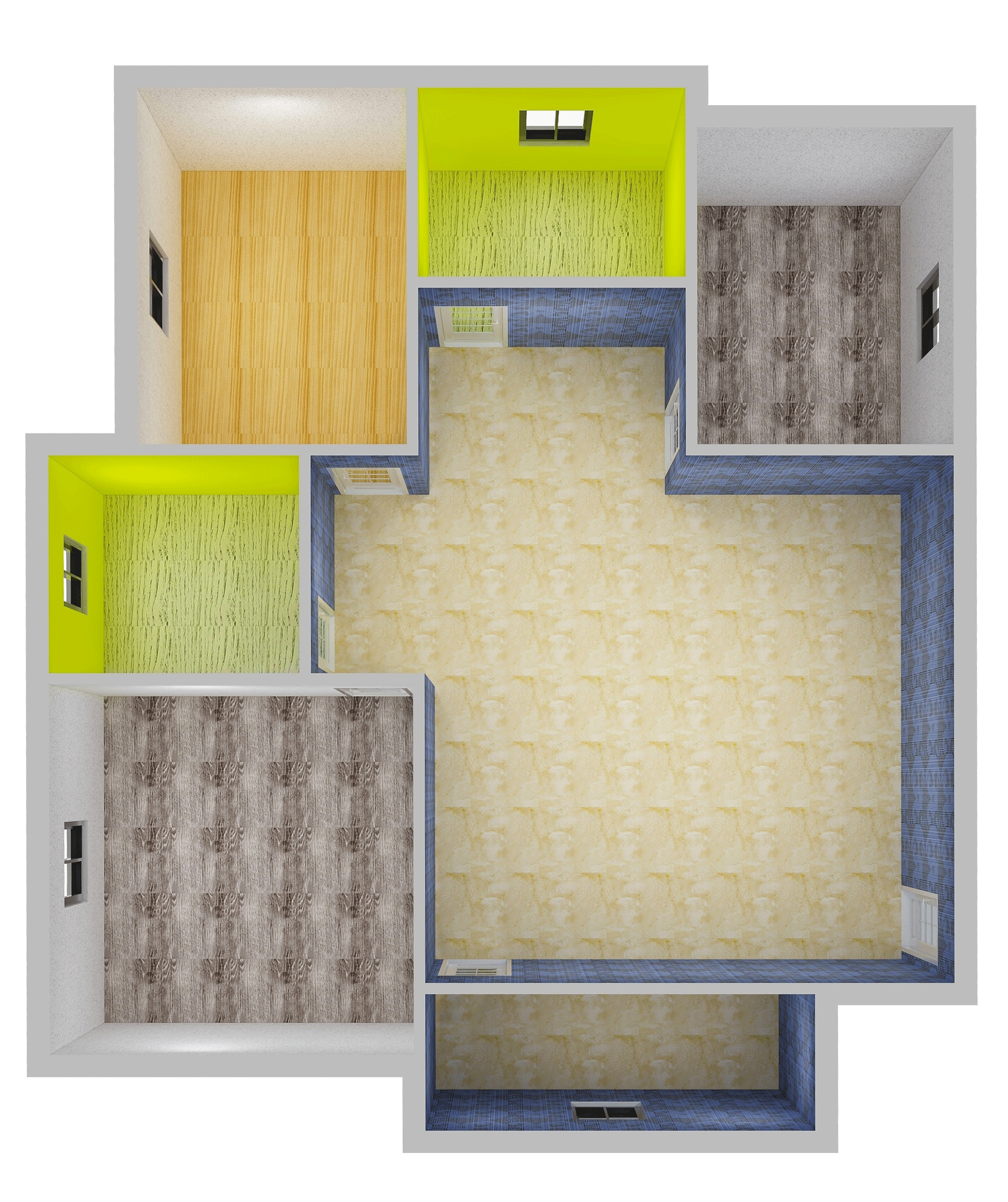} \\
				\hline
				\textit{The house plan has two bedrooms, one washroom, one balcony, one livingroom, and one kitchen. More specifically, bedroom1 has Pink Wall\_Cloth wall, and floor is Wood\_Veneer and White. bedroom1 covers 13 square meters located in north. In addition, bedroom2 has 11 squares in southwest. bedroom2 floor is White Wood\_Veneer as well as wall is Wall\_Cloth and Pink. Moreover, floor of washroom1 is Jade and Blue, and wall is White Wall\_Cloth. washroom1 covers 5 square meters located in south. Additionally, balcony1 has 4 squares in northeast. balcony1 uses Wood\_color Wood\_Grain for floor as well as wall is Wall\_Cloth and Earth\_color. In addition, livingroom1 covers 28 square meters located in east. floor of livingroom1 is Wood\_Grain and Wood\_color, and wall is Wall\_Cloth and Earth\_color. Additionally, kitchen1 has 5 squares in center. kitchen1 has Blue Jade floor as well as wall is White Wall\_Cloth. livingroom1 is adjacent to bedroom1, bedroom2, kitchen1, washroom1, balcony1. kitchen1 and bedroom1 are connected. bedroom2 is adjacent to kitchen1, washroom1.} 
				&\includegraphics[width=0.35\columnwidth]{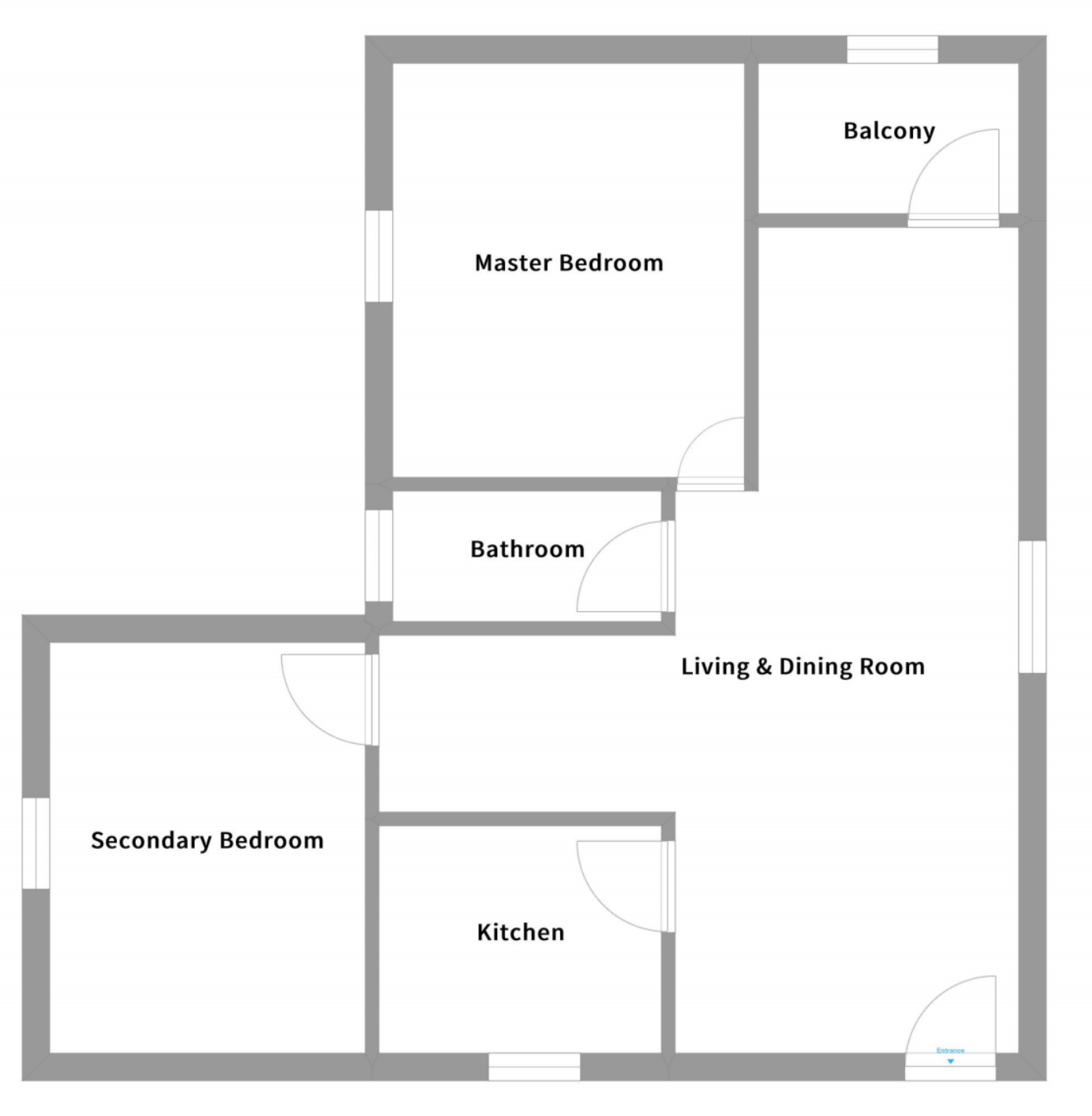} 
				&\includegraphics[width=0.35\columnwidth]{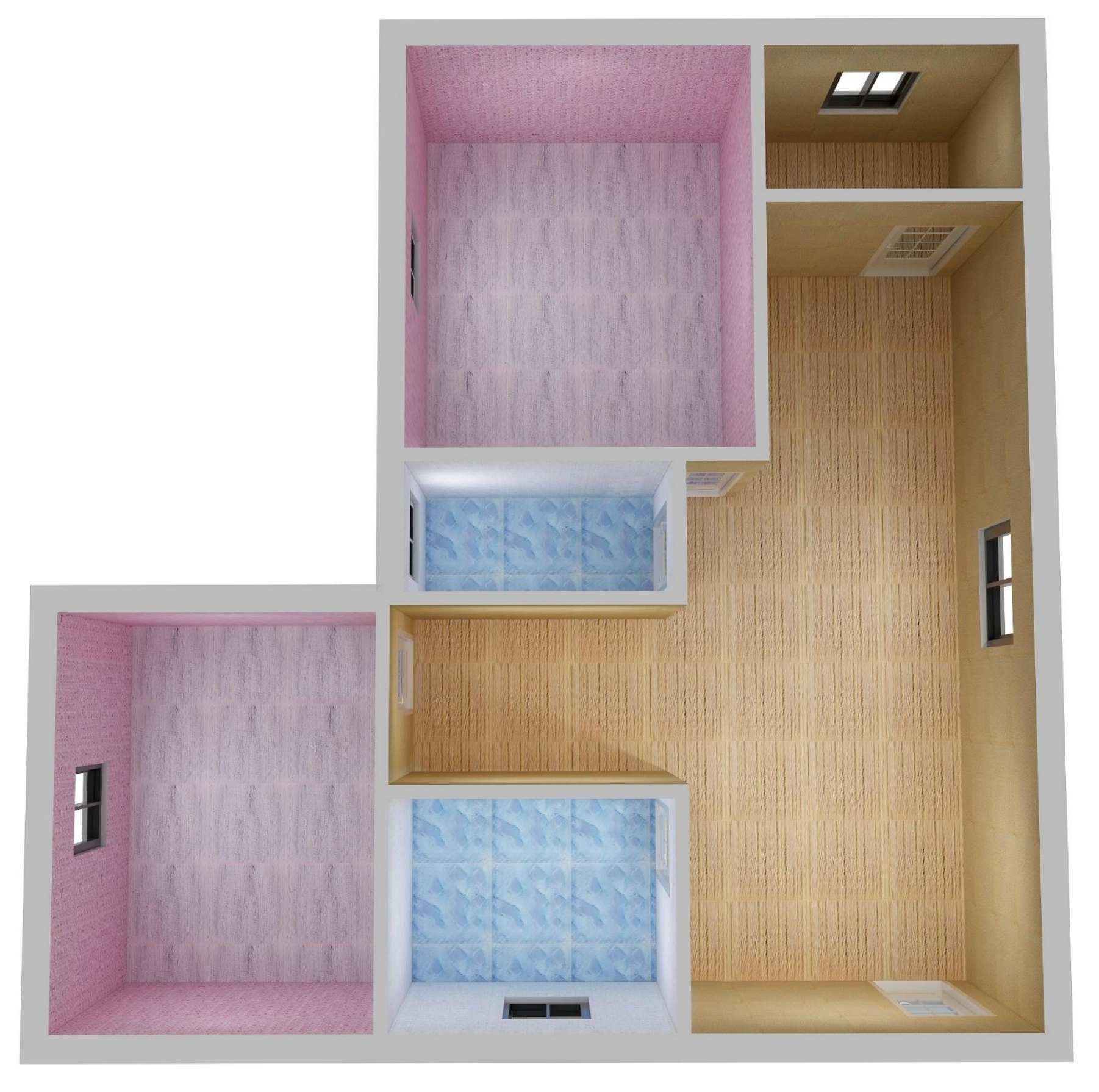}
				&\includegraphics[width=0.35\columnwidth]{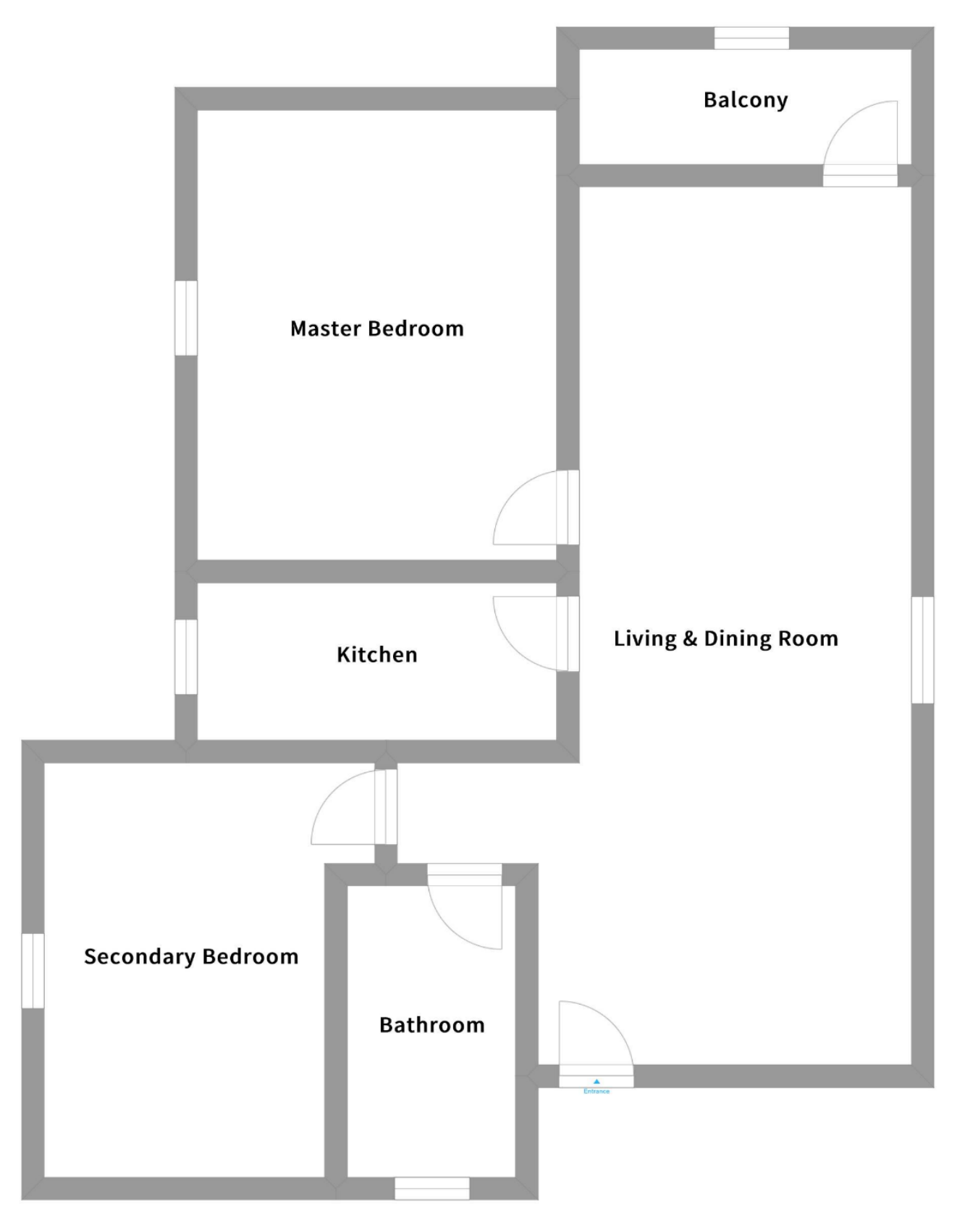} 
				&\includegraphics[width=0.35\columnwidth]{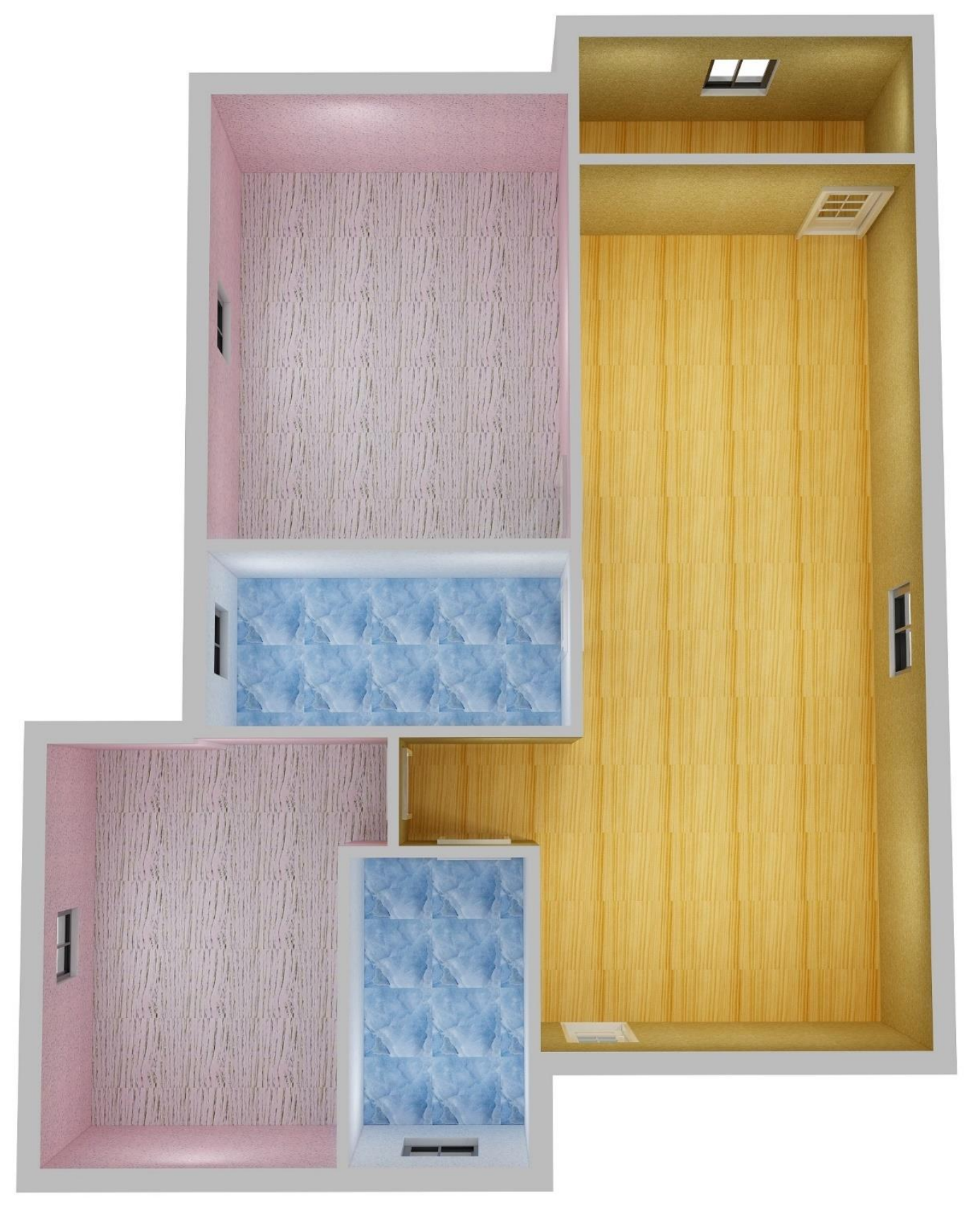}\\
				\hline
	\end{tabular}}}
	\vspace{5pt}
	\caption{More results of 3D house plans corresponding to the given input linguistic requirements.}
	\label{fig:more_3d_house_plan}
\end{figure*}

In this section, we report more visual results of 2D and 3D house plans corresponding to the given linguistic requirements, and show these results in Figure~\ref{fig:more_3d_house_plan}.

\end{document}